\documentclass[lettersize,journal]{IEEEtran}
\usepackage{amsmath,amsfonts}
\usepackage{algorithmic}
\usepackage{algorithm}
\usepackage{array}
\usepackage{textcomp}
\usepackage{stfloats}
\usepackage{url}
\usepackage{verbatim}
\usepackage{graphicx}
\usepackage{xcolor}
\usepackage{pifont}
\usepackage{multirow}
\usepackage{threeparttable}
\usepackage{enumitem}
\usepackage{cite}
\usepackage{bbding}
\usepackage{float} 
\usepackage{subfigure}
\usepackage{colortbl}
\begin{document}

\title{Event-Driven Learning for Spiking Neural Networks}

\author{Wenjie~Wei,
        Malu~Zhang,
        Jilin~Zhang,
        Ammar~Belatreche,
        Jibin Wu,
        Zijing~Xu,
        Xuerui~Qiu,
        Hong~Chen,
        Yang~Yang~and~Haizhou~Li,~\IEEEmembership{Fellow,~IEEE}
        % <-this % stops a space
\IEEEcompsocitemizethanks{
\IEEEcompsocthanksitem W. Wei, M. Zhang, Z. Xu, X. Qiu, Y. Yang are with the University of Electronic Science and Technology of China, Chengdu 610054, China. Email:~maluzhang@uestc.edu.cn%\protect\\% note need leading \protect in front of \\ to get a newline within \thanks as
\IEEEcompsocthanksitem J. Zhang and H. Chen are with the School of Integrated Circuits, Tsinghua University, Beijing, 100084, China. Email:~hongchen@tsinghua.edu.cn
\IEEEcompsocthanksitem A. Belatreche is with the Department of Computer and Information Sciences, Faculty of Engineering and Environment, Northumbria University, Newcastle upon Tyne NE1 8ST, U.K. Email:~ammar.belatreche@northumbria.ac.uk
\IEEEcompsocthanksitem J. Wu is with the Department of Computing, The Hong Kong Polytechnic University, 11 Yuk Choi Road, Hung Hom, KLN. Email:~jibin.wu@polyu.edu.hk
\IEEEcompsocthanksitem H. Li is with Shenzhen Research Institute of Big Data, School of Data Science, The Chinese University of Hong Kong, Shenzhen (CUHK-Shenzhen), China. Email:~haizhou.li@nus.edu.sg}}




\maketitle

\begin{abstract}
Brain-inspired spiking neural networks (SNNs) have gained prominence in the field of neuromorphic computing owing to their low energy consumption during feedforward inference on neuromorphic hardware. However, it remains an open challenge how to effectively benefit from the sparse event-driven property of SNNs to minimize backpropagation learning costs. In this paper, we conduct a comprehensive examination of the existing event-driven learning algorithms, reveal their limitations, and propose novel solutions to overcome them. Specifically, we introduce two novel event-driven learning methods: the spike-timing-dependent event-driven (STD-ED) and membrane-potential-dependent event-driven (MPD-ED) algorithms. These proposed algorithms leverage precise neuronal spike timing and membrane potential, respectively, for effective learning. The two methods are extensively evaluated on static and neuromorphic datasets to confirm their superior performance. They outperform existing event-driven counterparts by up to 2.51\% for STD-ED and 6.79\% for MPD-ED on the CIFAR-100 dataset. In addition, we theoretically and experimentally validate the energy efficiency of our methods on neuromorphic hardware. On-chip learning experiments achieved a remarkable 30-fold reduction in energy consumption over time-step-based surrogate gradient methods. The demonstrated efficiency and efficacy of the proposed event-driven learning methods emphasize their potential to significantly advance the fields of neuromorphic computing, offering promising avenues for energy-efficiency applications.

 %we provide an in-depth theoretical analysis of the training complexity of our methods and conduct additional validation experiments. Results demonstrate that our methods could achieve nearly identical learning effects to the surrogate gradient (SG) learning algorithm but with lower computational costs. Finally, we deploy our method on a recently developed neuromorphic chip, and on-chip learning results indicate that ours achieves 29$\times$ energy savings compared to the SG method.
%Also, this is the first time that an event-driven learning algorithm can successfully train the ImageNet dataset with competitive accuracy compared with deep learning. 

\end{abstract}


\section{Introduction}
\IEEEPARstart{D}{eep} Neural Networks (DNNs) have demonstrated remarkable success in many applications, such as computer vision, speech recognition, and natural language processing~\cite{alzubaidi2021review,dong2021survey}. However, DNNs generally rely on the availability of ample computing resources, that limits their applications on power-critical computing platforms, such as edge computing~\cite{feldmann2019all,deng2020rethinking}. Inspired by brain computing, Spiking Neural Networks (SNNs) are proposed to offer an ultra-low-power alternative for DNNs~\cite{izhikevich2003simple,gerstner2002spiking,roy2019towards}. SNNs encode information by binary spikes over time and work in a sparse event-driven manner, which not only gives rise to the potential of powerful spatiotemporal information processing capabilities but also enables the deployment on ultra-low power neuromorphic hardware~\cite{akopyan2015truenorth,pei2019towards,davies2018loihi}, such as recently developed TrueNorth~\cite{akopyan2015truenorth}, Tianjic~\cite{pei2019towards}, and Loihi~\cite{davies2018loihi}.

However, unlike DNNs which have mature backpropagation (BP) algorithms as the workhorse of learning, it remains a challenge how learning algorithms effectively benefit from the sparse event driven property of  SNNs due to the complex spatiotemporal neuronal dynamics and the non-differentiability nature of discrete spike events~\cite{zhang2021rectified,wei2023temporal,zhang2019spike,zhang2023self}. As such, there remains a performance gap between SNNs and their DNN counterparts when applied to a wide range of challenging real-world tasks.

In order to overcome the challenges in training deep SNNs, several studies convert a pretrained high-performance DNN to its corresponding SNN version~\cite{rueckauer2016theory, rueckauer2018conversion,han2020rmp,wang2022signed}. Despite many successes, these ANN-to-SNN conversion methods significantly increase the inference latency and are unsuitable for processing spatiotemporal neuromorphic data. Surrogate gradient (SG) learning methods are proposed to directly train deep SNNs with the ability to efficiently process spatiotemporal data~\cite{neftci2019surrogate, wu2018spatio,wu2019direct,shrestha2018slayer}. However, the gradient in SG methods is propagated at each time step, resulting in a substantial increase in time complexity and memory usage~\cite{zhu2022training,zhu2023exploring,Yin2023MINTMI}.

Unlike ANN-to-SNN and SG methods, event-driven algorithms train SNNs in response to specific events or spikes generated by neurons in the SNN, holding substantial potential to significantly reduce memory usage and computational costs during the learning process~\cite{zhang2020temporal,zhu2023exploring}. As a result, SNNs trained with event-driven methods exhibit advantages in both training and inference when deployed on ultra-low power neuromorphic hardware. However, existing event-driven methods are underdeveloped compared to SG methods. This work aims to bridge this gap by proposing simple yet effective and efficient event-driven learning algorithms for deep SNNs. The main contributions of this work are summarized as follows:
\begin{itemize}
\item We examine the challenges associated with training SNNs in an event-driven manner, addressing issues such as over-sparsity and gradient reversal. Furthermore, we conduct a comprehensive analysis of the limitations in existing approaches aimed at overcoming these challenges.

\item  We propose the Spike-Timing-Dependent Event-Driven (STD-ED) learning algorithm for deep SNNs. We first introduce the Adaptive Firing Threshold-based Integrate-and-Fire (AFT-IF) neuron to address the problems of over-sparsity and gradient reversal. Based on this AFT-IF neuron, we then present the STD-ED algorithm for SNNs where learning occurs only at spike times, following a fully event-driven approach.

\item We propose the Membrane-Potential-Dependent Event-Driven (MPD-ED) learning algorithm for deep SNNs. This algorithm integrates the proposed AFT mechanism into spiking neurons and introduces the masked surrogate gradient function to implement the MPD-ED approach. In MPD-ED, the membrane potential is used as the learning signal, and training occurs only when the membrane potential exceeds the firing threshold.

\item Extensive experiments are conducted on both static and neuromorphic datasets. The obtained results demonstrate that our proposed methods achieve state-of-the-art performance when compared with other existing event-driven approaches. Furthermore, we validate the energy efficiency of our methods through theoretical analysis and hardware implementation. On-chip learning experiments reveal a remarkable 30-fold reduction in energy consumption compared to time-step-based SG counterparts.
\end{itemize}
The rest of the paper is organized as follows.
In Section 2, we provide a comprehensive review of existing learning algorithms for SNNs, containing ANN-to-SNN conversion methods, surrogate gradient methods, and event-driven methods.
In Section 3, we present the preliminaries and analyze the challenges associated with training SNNs in an event-driven manner.
In Section 4, we introduce two proposed event-driven learning algorithms, namely STD-ED and MPD-ED, and provide a thorough analysis and summary of these two algorithms.
In Section 5, we evaluate the performance of our methods on multiple benchmark datasets using various network architectures. Additionally, we conduct ablation studies to prove the effectiveness of the crucial components in the two proposed algorithms.
In Section 6, we validate the energy efficiency and practical applicability of our methods through theoretical analysis and hardware implementation.
Finally, we conclude the paper in Section 7.

\section{Related Work}
In order to effectively train deep SNNs, various algorithms have been proposed, which can be broadly categorized into three groups: ANN-to-SNN conversion methods, surrogate gradient methods, and event-driven methods.

\subsection{ANN-to-SNN conversion} These methods avoid the learning difficulties of SNNs by first training a high-performance ANN and subsequently converting it to an SNN version. This type of method benefits from the mature learning algorithm of ANNs while facing the trade-off problem between accuracy and inference latency. To enable a converted SNN with high performance and less inference latency, various strategies have been proposed such as normalization~\cite{diehl2015fast,rueckauer2016theory,rueckauer2017conversion}, threshold balancing~\cite{diehl2015fast,diehl2016conversion,sengupta2019going}, the soft-reset mechanism~\cite{rueckauer2017conversion,han2020rmp}, optimized potential initialization~\cite{bu2022optimized} and layer-wise calibration~\cite{li2021free,wu2021tandem,wu2021progressive,bu2023optimal}. Despite the excellent accuracy and reduced inference time steps in recent literature~\cite{wang2022signed,hao2023reducing,hao2023reducing1}, these conversion methods utilize the rate-based coding scheme, leaving room for further enhancement in energy efficiency. A few works explore the ANN-to-SNN method with the temporal coding scheme~\cite{rueckauer2018conversion,stanojevic2022exact}, which conveys information through precise spike timing~\cite{park2020t2fsnn} or the time difference between two spikes~\cite{han2020deep}. While the converted SNN with temporal coding improves energy efficiency, it suffers from severe performance degradation over short time steps. Moreover, the ANN-to-SNN conversion method cannot process spatiotemporal neuromorphic data, resulting in the powerful spatiotemporal information processing capability of SNNs not being fully exploited.

\subsection{Surrogate gradient learning} In this area, SNNs are treated as binary-output Recurrent Neural Networks (RNNs), with the discontinuities of binary spikes being handled via surrogate gradients, which draw inspiration from the backpropagation through time (BPTT) algorithm~\cite{neftci2019surrogate,wu2018spatio,wu2019direct,shrestha2018slayer,zenke2018superspike,lee2020enabling, qiu2023vtsnn, qiu2023gated}. Compared with ANN-to-SNN conversion methods, SG methods offer a direct training approach for SNNs, yielding higher accuracy with reduced inference latency for both static and neuromorphic datasets. The performance of SG methods is further enhanced by introducing parametric spiking neurons~\cite{fang2021incorporating}, more suitable surrogate functions~\cite{li2021differentiable,chen2022gradual}, and more efficient loss functions~\cite{deng2022temporal}. Although competitive performance has been reported on challenging datasets, such as CIFAR and ImageNet, the gradient descent in SG methods is not well aligned with the real loss landscape of SNNs, and the learning process is susceptible to obtaining a locally optimal solution with limited generalizability~\cite{deng2022temporal}. Moreover, the gradient information in SG is propagated at each time step, leading to high computational costs and memory usage when performing training~\cite{xiao2022online}. Recently, several studies on SG learning have emerged to reduce training demands~\cite{perez2021sparse,yang2022training,xiao2022online,meng2023towards} or have been applied to large network structure~\cite{yao2023attention,zhu2022tcja}, but none of them exploit the sparse event-driven nature of SNN in the backward propagation process.

\subsection{Event-driven learning}
Event-driven algorithms train SNNs in an event-driven manner and regard precise spike timing as a relevant signal for synapse updating. SpikeProp~\cite{bohte2002error} and its variants~\cite{zhao2018convergence,mckennoch2006fast} are the pioneers in this field. By assuming the membrane potential increases linearly around the spike time, these methods can successfully calculate the derivative of spike timing to membrane potential. The performance of SpikeProp-based methods is further improved by applying non-leaky spiking neuron models~\cite{mostafa2017supervised,kheradpisheh2020temporal,comcsa2021temporal}, such as Integrate-and-fire (IF) neurons~\cite{kheradpisheh2020temporal} and ReL-PSP neurons~\cite{zhang2021rectified}. However, these methods are based on the time-to-first-spike (TTFS) coding scheme where each neuron is constrained to fire at most once, so they cannot be applied to process data with multiple spikes, i.e., neuromorphic datasets. TSSL-BP~\cite{zhang2020temporal} is proposed to train deep SNNs with the rate-based coding scheme. However, it requires the help of surrogate gradient learning and cannot work in a purely event-driven manner.  Recently, Zhu et al.~\cite{zhu2022training,zhu2023exploring} propose a purely event-driven learning algorithm for SNNs. As they apply a smoother gradient function to address the gradient reversal problem, the feedforward and backward of SNNs are inconsistent and the accuracy performance could be improved. Moreover, to overcome the over-sparsity problem, Zhu et al.~\cite{zhu2022training,zhu2023exploring} utilize a binary search to determine suitable initialization parameters that guarantee each layer's average firing rate reaches a specified value. Nonetheless, this method is not only time-intensive but also incapable of addressing gradient-blocking issues in the learning phase.

%where they first discover and address the gradient reversal problem. However, Zhu et al.~\cite{zhu2022training,zhu2023exploring} address the over-sparsity problem by employing a binary search method, which determines suitable initialization parameters to guarantee that the average spike activity of each layer reaches a specified value. However, this method is not only time-intensive but also cannot address the over-sparsity issue in the learning process.

 %in this method, the forward and backward of SNNs are inconsistent, and the performance can be further improved. 
%Recently, Zhu et al.~\cite{zhu2023exploring} further improve their work by replacing the original mean square counting loss with an enhanced counting loss and transferring the training of scale factor in weight standardization into threshold. Although the performance is further improved compared to the previous work, they still use the smooth gradient function, and there is still a lot of room for performance improvement.

\section{Preliminary and Problem Analysis}

In this section, we first provide an overview of the classical event-driven learning algorithms. Then, we delve into the challenges associated with training SNNs in an event-driven manner and analyze the limitations in existing approaches aimed at overcoming these challenges.

\begin{figure*}
\centering
\subfigure[]{
\includegraphics[scale=0.56]{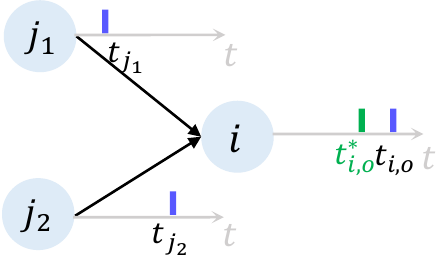}}
% \hspace{0.015\linewidth}
\subfigure[]{
\includegraphics[scale=0.54]{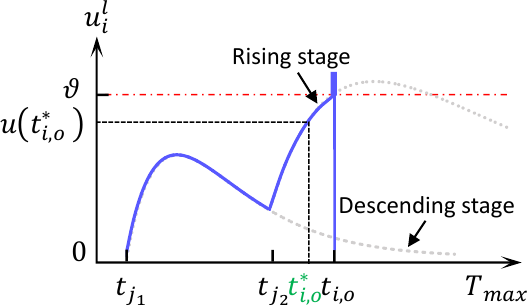}}
\subfigure[]{
\includegraphics[scale=0.54]{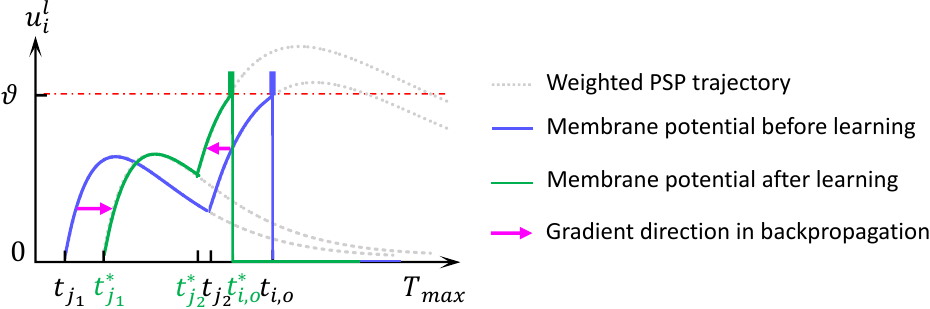}}
\caption{A case of the gradient reversal problem. (a) Neuron $i$ is connected to two afferent neurons $j_1$ and $j_2$ with positive synaptic weights. It generates a spike at time $t_{i,o}$, while we want it to fire earlier at $t_{i,o}^*$. (b) The spike $t_{i,o}$ is generated in the descending stage of PSP induced by $t_{j_1}$ and the rising stage of PSP induced by $t_{j_2}$. (c) To make neuron $i$ fire earlier towards $t_{i,o}^*$, the derivative of $\partial \mathcal{L}/\partial t_{i,o}$ should be positive according to the stochastic gradient descent rule, i.e., $t_{i,o}^*=t_{i,o}-\gamma \frac{\partial \mathcal{L}}{\partial t_{i,o}}$. In the backpropagation-based algorithm, the derivatives of $\partial \mathcal{L}/\partial t_{j_1}$ and $\partial \mathcal{L}/\partial t_{j_2}$ are expected to be positive due to the positive synaptic connections. However, due to the alpha-shaped kernel, the event-driven learning of SNNs is not well-suited for the backpropagation-based learning algorithm. In the event-driven learning process, achieving an early spike generation for neuron $i$ at $t_{i,o}^*$ requires an increase in the membrane potential $u(t_{i,o}^*)$, which involves enhancing the contributions of $t_{j_1}$ and $t_{j_2}$ to the membrane potential of neuron $i$. The event-driven learning of SNNs achieves this by guiding $t_{j_1}$ to occur later at $t_{j_1}^*$ and $t_{j_2}$ to occur earlier at $t_{j_2 }^*$, resulting in a positive derivative of $\partial \mathcal{L}/\partial t_{j_2}$ and a negative derivative of $\partial \mathcal{L}/\partial t_{j_1}$, where the derivative of $\partial \mathcal{L}/\partial t_{j_1}$ is not behave as expected.}
\label{fig:reverse} 
\end{figure*}
% a positive derivative of $\partial \mathcal{L}/\partial t_{j_2}$ to guide $t_{j_2}$ to occur earlier at $t_{j_2 }^*$, and a negative derivative of $\partial \mathcal{L}/\partial t_{j_1}$ to guide $t_{j_1}$ to occur later at $t_{j_1}^*$

\subsection{Preliminary}

Among various spiking neuron models that simulate the information processing capability of biological neurons~\cite{hodgkin1952quantitative,gerstner2002spiking}, the Spike Response Model~(SRM)~\cite{gerstner1992associative} is the most widely used in existing event-driven learning algorithms~\cite{zhang2020temporal,xu2013supervised,zhu2022training}. The membrane potential of an SRM neuron $i$ in the $l$-th layer is defined as
\begin{equation}
u^l_i(t)=\sum_{j}\sum_{t_{j,f}^{l-1}}\omega^l_{ij}K(t-t^{l-1}_{j,f})-\sum_{t_{i,k}^l}\eta(t-t_{i,k}^l)+u_{rest},
\label{srmneuron}
\end{equation}
where $u_{rest}$ denotes the resting potential in the absence of input spikes. The first term in Eq.~\ref{srmneuron} denotes the integrated input current, where $t_{j,f}^{l-1}$ is the time of the $f$-th spike from afferent neuron $j$ in layer $l-1$, and $\omega_{ij}^l$ is the synaptic weight between neuron $j$ and neuron $i$. Each incoming spike will induce a postsynaptic potential (PSP), commonly defined as
\begin{equation}
   K(t)=\frac{\tau_m}{\tau_m-\tau_s} \left [ \exp\left(\frac{-t}{\tau_m}\right)-\exp\left(\frac{-t}{\tau_s}\right) \right ], ~~~t>0.
   \label{alphaPSP}
\end{equation}
The induced PSP exhibits an alpha shape, which is controlled by membrane time constant $\tau_m$ and synapse time constant $\tau_s$. Neuron $i$ integrates weighted PSPs and emits spikes when the firing condition is satisfied, i.e., the membrane potential exceeds the threshold $\theta$. Mathematically, $\mathcal{F}_i^l$ is a set of spike timings satisfying the firing condition, which is stated as
\begin{equation}
% s^l_i(t)=\mathcal{H}(u^l_i(t)-\vartheta),
\mathcal{F}_i^l = \left \{t_{i,k}^l | u_i^l(t_{i,k}^l) \geq \theta \right \},
\label{spiketrain}
\end{equation}
where $k\in\mathbb{N}^+$ is an index of the spike. The second term in Eq.~\ref{srmneuron} is the refractory function, depicting the response of the membrane potential to output spikes. The refractory kernel $\eta$ is typically defined as
\begin{equation}
\eta(t)=\theta exp\left ( \frac{-t}{\tau_m}\right), ~~~t>0.
\end{equation}

To implement event-driven learning, it is necessary to calculate the derivatives of the output spike with respect to synaptic weight and input spike. According to Eq.~\ref{srmneuron} and Eq.~\ref{spiketrain}, we can get these two derivatives as follows
\begin{equation}
   \frac {\partial t_{i,k}^{l}} {\partial \omega^l_{ij}}= \frac {\partial t_{i,k}^{l}} {\partial u_i^l(t^l_{i,k})}
   \frac  {\partial u_i^l(t^l_{i,k})}{\partial \omega^l_{ij}},~~~\frac {\partial t_{i,k}^{l}} {\partial t_{j,f}^{l-1}}=\frac {\partial t_{i,k}^{l}} {\partial u_i^l(t^l_{i,k})}\frac  {\partial u_i^l(t^l_{i,k})}{\partial t_{j,f}^{l-1}}.
   \label{tm/w}
\end{equation}
% \begin{equation}
%    \frac {\partial t_{i,k}^{l}} {\partial t_{j,f}^{l-1}}=\frac {\partial t_{i,k}^{l}} {\partial u_i^l(t^l_{i,k})}\frac  {\partial u_i^l(t^l_{i,k})}{\partial t_{j,f}^{l-1}}.
%    \label{tm/tk}
% \end{equation}
where the derivatives of the membrane potential to synaptic weight and input spike can be easily obtained. However, the calculation of $\partial t_{i,k}^{l}/\partial u_i^l(t^l_{i,k})$ poses a challenge due to the spike generation function. Existing event-driven approaches replace this computation with $-(\partial u_i^l(t^l_{i,k})/\partial t_{i,k}^{l})^{-1}$~\cite{bohte2002error,zhang2021rectified,zhu2023exploring}. Consequently, the SNN can be trained successfully in an event-driven manner.

\subsection{Problem analysis}

\subsubsection{Over-sparsity problem}

Due to the inherent leaky properties of membrane potential and the spike generation mechanism, deep SNNs suffer from the over-sparsity problem~\cite{lee2016training,guo2022loss}. This problem becomes especially serious in event-driven learning algorithms, as the gradient is only propagated through the generated spike. Mathematically, as shown in Eq.~\ref{tm/w}, if neuron $i$ fails to generate a spike at time $t_{i,k}^l$, the error cannot be backpropagated via ${\partial t_{i,k}^l}/{\partial u_i^l(t_{i,k}^l)}$.  In an extreme scenario where one layer in SNNs fails to generate any spike after initialization, the gradient information is completely blocked by this layer, making the overall training impossible. Therefore, maintaining a certain number of active neurons in each layer is crucial during the event-driven learning process.

In order to address the over-sparsity problem that hinders event-driven learning, Zhang et al.~\cite{zhang2021rectified} and Wei et al.~\cite{wei2023temporal} propose a linearly increased PSP function and a linearly decreased firing threshold, respectively. However, these methods impose a constraint on the spiking neuron to fire at most once, making them unsuitable for processing sequence data. Zhu et al.~\cite{zhu2022training,zhu2023exploring} employ a binary search technique to identify appropriate initialization parameters, ensuring that the average spike activity of each layer reaches a predetermined value. However, this method is not only time-intensive in the initialization stage but also incapable of addressing over-sparsity issues in the learning phase.

\subsubsection{Gradient reversal problem}
\label{gradient_reversal}
The gradient reversal problem arises from the mismatch between the backpropagation-based learning algorithm and spiking neurons. As shown in Fig.~\ref{fig:reverse}, neuron $i$ is connected to two presynaptic neurons with positive synaptic weights. The spike $t_{i,o}$ occurs in the descending stage of PSP induced by $t_{j_1}$ (i.e., $dK(t-t_{j_1})/dt|_{t=t_{i,o}}<0$) and in the rising stage of PSP induced by $t_{j_2}$ (i.e., $dK(t-t_{j_2})/dt|_{t=t_{i,o}}>0$). To achieve an early spike for neuron $i$, the derivative of $\partial \mathcal{L}/\partial t_{i,o}$ should be positive. Moreover, in the backpropagation algorithm, the derivatives of $\partial \mathcal{L}/\partial t_{j_1}$ and $\partial \mathcal{L}/\partial t_{j_2}$ should also be positive due to positive synaptic connections. However, due to the alpha-shaped kernel of spiking neurons, the event-driven learning of SNNs is not well-suited for the traditional backpropagation-based learning algorithm. As shown in Fig.~\ref{fig:reverse}(c), to make neuron $i$ fire an earlier spike at $t_{i,o}^*$, the event-driven learning of SNNs requires a later $t_{j_1}$ and an earlier $t_{j_2}$ to contribute more PSP to the membrane potential of neuron $i$ at $t_{i,o}^*$. This results in a positive $\partial \mathcal{L}/\partial t_{j_2}$ and a negative $\partial \mathcal{L}/\partial t_{j_1}$, where the gradient on $t_{j_1}$ is reversed. In summary, due to the alpha-shape PSP kernel, the gradients in event-driven learning may not behave as expected in the backpropagation-based algorithm. This gradient reversal phenomenon results in an unstable training process and slow convergence, hindering event-driven algorithms from keeping pace with well-developed SG algorithms.

Early event-driven approaches neglect the gradient reversal problem, but they propose several strategies that indirectly address and mitigate this issue.
For instance, Zhang et al.~\cite{zhang2021rectified} introduce a ReL-PSP neuron model, and Zhang et al.~\cite{zhang2020temporal} utilize a sigma function to assist training. 
In~\cite{zhu2022training}, Zhu et al. point out the gradient reversal problem and address it by substituting the derivative of $dK(t)/dt$ with a continuous and positive function $h(t)=e^{-t/\tau_{grad}}$ during backpropagation. However, this modification introduces a mismatch between the feedforward computation and backpropagation learning, potentially impacting the accuracy performance. Recently, in their follow-up work ~\cite{zhu2023exploring}, Zhu et al. further refine their approach by incorporating an improved counting loss. However, they still employ the alpha-shaped kernel, which still suffers from the inconsistency problem and the accuracy could be improved.

\section{Methods}
In this section, we introduce two proposed event-driven learning algorithms, namely STD-ED and MPD-ED. Furthermore, we provide a thorough analysis and summary of both approaches.

\subsection{Spike-timing-dependent event-driven learning algorithm}

\begin{figure}
\centering
\includegraphics[scale=0.55]{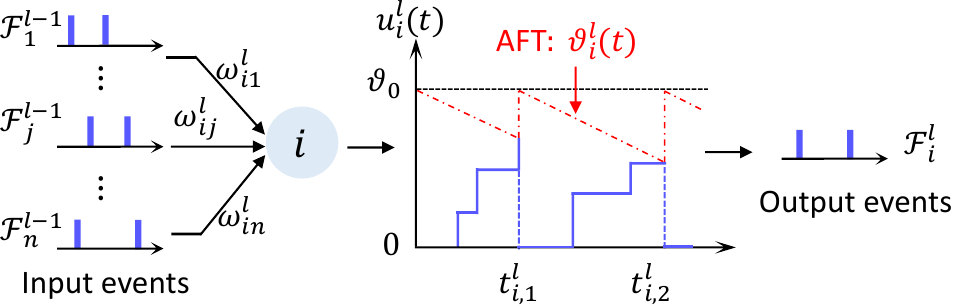}
\caption{Adaptive firing threshold-based Integrate-and-Fire (AFT-IF) neuron.}
\label{fig:AFTIF}
\end{figure}

\subsubsection{Neuron model} 
To resolve the problems of over-sparsity and gradient reversal, we first propose a novel Adaptive Firing Threshold-based Integrate-and-Fire (AFT-IF) spiking neuron model. The membrane potential of an AFT-IF neuron, induced by presynaptic neurons, can be mathematically described as
\begin{equation}
u^l_i(t)=\sum_{j}\sum_{t_{j,f}^{l-1}}\omega^l_{ij}K(t-t^{l-1}_{j,f}),
% -\underbrace{\sum_{t_{i,k}^l<t}\eta(t-t_{i,k}^l)}_{\text{intra-neuron dynamics}}
\label{AFTIF}
\end{equation}
where the $K(\cdot)$ is the PSP function and is defined as 
\begin{equation}
K(t-t^{l-1}_{j,f}) = \left\{
\begin{array}{ll}
1, & {t_{i,last}^{l} \leq t^{l-1}_{j,f} < t},\\
0, & \text{otherwise},
\end{array} \right.
\label{nonleakyPSP}
\end{equation}
where $t_{i,last}^{l}$ is the latest output spike of neuron $i$. As shown in Fig.~\ref{fig:AFTIF}, the PSP kernel $K(\cdot)$ is characterized by its non-leaky property. In addition, the AFT-IF neuron is also distinguished by its adaptive firing threshold (AFT) mechanism, which is defined as 
\begin{equation}
\vartheta_i^l(t) = \left\{
\begin{array}{ll}
\vartheta_0-\alpha(t-t_{i,last}^{l}), &~  t_{i,last}^{l}<t,\\
\vartheta_0, &~ t_{i,last}^{l}=t,
\end{array} \right.
\label{AFT}
\end{equation}
where $\vartheta_0$ is the initial threshold, it is set to 1 in our work. As shown in Fig.~\ref{fig:AFTIF}, the time-varying AFT undergoes decay with the parameter $\alpha$, while it increases to the initial value $\vartheta_0$ after each spike emission. Note that the threshold $\vartheta_i^l(t)$ will not fall below 0. The AFT-IF neuron $i$ emits a spike when its membrane potential $u_i^l(t)$ reaches the firing threshold $\vartheta_i^l(t)$. Therefore, the spike times of neuron $i$ is defined as the set of
\begin{equation}
\mathcal{F}_i^l = \left \{t_{i,k}^{l} | u_i^l(t_{i,k}^{l}) \geq \vartheta_i^l(t_{i,k}^{l})\right\},
\label{spikefunction}
\end{equation}
where $k$ is the index of the spike. After each spike emission, the hard reset mechanism is employed, where the membrane potential is reset to 0.

\subsubsection{STD-ED learning rule}

In order to train SNNs with the time of spikes in an event-driven manner, it is necessary to compute derivatives of the loss function with respect to input spikes and synaptic weights, i.e., ${\partial{\mathcal L}}/{\partial{t_{j,f}^{l-1}}}$ and ${\partial{\mathcal L}}/ {\partial\omega_{ij}^{l}}$. In the following, we will describe how the proposed STD-ED resolves these two terms.

We begin by resolving ${\partial{\mathcal L}}/{\partial{t_{j,f}^{l-1}}}$. The input spike $t_{j,f}^{l-1}$ influences the loss function ${\mathcal L}$ by affecting the output spike of neuron $i$ in the $l$-th layer, i.e., $t_{i,k}^l$, so the derivative can be expressed as
\begin{equation}
\frac{\partial{\mathcal L}}{\partial t_{j,f}^{l-1}}=\sum_{i}\frac {\partial{\mathcal L}}{\partial t_{i,k}^{l}}\frac {\partial t_{i,k}^{l}}{\partial t_{j,f}^{l-1}}.
% =\sum_{i}\delta_{i,k}^{l}\frac {\partial t_{i,k}^{l}}{\partial t_{j,f}^{l-1}},
\label{l/t}
\end{equation}
The first item of Eq. \ref{l/t} is the derivative of the loss function to the output spike, and it can be computed recursively. The second item represents the derivative between layers, where the input spike $t_{j,f}^{l-1}$ influences the output spike $t_{i,k}^l$ by affecting the membrane potential of neuron $i$. Therefore, the derivative between layers can be further calculated as
\begin{equation}
\frac {\partial t_{i,k}^{l}}{\partial t_{j,f}^{l-1}}
=\frac{\partial t_{i,k}^l}{\partial u_i^l(t_{i,k}^l)}\frac {\partial u_i^l(t_{i,k}^l)}{\partial t_{j,f}^{l-1}}.
\label{t/t}
\end{equation}
To compute the derivative of ${\partial u_i^l(t_{i,k}^l)}/{\partial t_{j,f}^{l-1}}$, we should note that reducing input $t_{j,f}^{l-1}$ will increase the membrane potential $u_i^l(t_{i,k}^l)$ by $\omega_{ij}^l$ earlier in time, hence we approximate ${\partial u_i^l(t_{i,k}^l)}/{\partial t_{j,f}^{l-1}}=-\omega_{ij}^l$~\cite{kheradpisheh2020temporal}. Furthermore, we compute the derivative of ${\partial t_{i,k}^l}/{\partial u_i^l(t_{i,k}^l)}$ by adopting the linear assumption~\cite{bohte2002error} that the membrane potential increases linearly in the infinitesimal time interval surrounding the spike time, thus
\begin{equation}
\begin{split}
\frac{\partial{t_{i,k}^l}}{\partial u_i^l(t_{i,k}^l)}
=\!-\left(\frac{\partial u_i^l(t_{i,k}^l)}{\partial{t_{i,k}^l}}\right)^{-1}\!=\!-{\left( \sum_{j}\sum_{t_{j,f}^{l\!-\!1}\in\mathcal{C}} \omega_{ij}^{l}\right)}^{-1}\!\!,
\end{split}
\label{t/u}
\end{equation}
where $\mathcal{C}$ represents a set of input spikes that contribute to the firing of $t_{i,k}^l$. Finally, in conjunction with Eqs.[\ref{l/t}-\ref{t/u}], the derivative of the loss function to the input spike becomes attainable.

We now derive the calculation of ${\partial{\mathcal L}}/{\partial\omega_{ij}^{l}}$. The synaptic weight $\omega_{ij}^{l}$ influences the loss function ${\mathcal L}$ by affecting the membrane potential and further the output spike of neuron $i$, so the derivative can be decomposed into the following equation through the chain rule
\begin{equation}
\frac {\partial{\mathcal L}}{\partial\omega_{ij}^{l}}
=\sum_{t_{i,k}^l}\frac{\partial \mathcal{L}}{\partial t_{i,k}^l}\frac{\partial t_{i,k}^l}{\partial u_i^l(t_{i,k}^l)}\frac{\partial u_i^l(t_{i,k}^l)}{\partial \omega_{ij}^l}.
\label{l/w}
\end{equation}
The first item is the derivative of the loss function to output spike, and the calculation of it has been described in Eq.~\ref{l/t}. The second and the third items can be determined through Eq.~\ref{t/u} and Eq.~\ref{AFTIF}, respectively. Therefore, the derivative of ${\partial{\mathcal L}}/{\partial\omega_{ij}^{l}}$ can be summarized as follows
\begin{equation}
\begin{split}
\frac {\partial{\mathcal L}}{\partial\omega_{ij}^{l}}\!=\!-\!\sum_{t_{i,k}^l}\frac{\partial \mathcal{L}}{\partial t_{i,k}^l}{\left(\! \sum_{j}\sum_{t_{j,f}^{l-1}\in\mathcal{C}} \omega_{ij}^{l}\!\right)\!}^{-1}\!\!\left(\!\sum_{t_{j,f}^{l-1}}K(t_{i,k}^l-t_{j,f}^{l-1})\!\right)\!.
\end{split}
\label{s:l/w}
\end{equation}
Consequently, the synaptic weight can be updated using the stochastic gradient descent method, i.e., $\omega_{ij}^l=\omega_{ij}^l-\gamma \frac{\partial{\mathcal L}}{\partial\omega_{ij}^{l}}$, where $\gamma$ is the learning rate parameter. In summary, both Eq.~\ref{l/t} and Eq.~\ref{s:l/w} constitute the gradient backpropagation formulas, allowing the SNN to be trained successfully with precise spike timings through the STD-ED method.

\subsubsection{Analysis and summary}The STD-ED algorithm incorporates the AFT-IF neuron that exhibits two features, including the IF kernel and the AFT mechanism. These two features effectively resolve the issues of gradient reversal and over-sparsity spikes. As analyzed earlier, the gradient reversal problem arises from the negative value of $dK(t)/dt$ in the backpropagation~\cite{zhu2022training}. In our approach, we employ the IF kernel to ensure that the PSP does not decay over time. As a result, the issue of gradient reversal is circumvented. In addition, the over-sparsity problem is addressed from two aspects. On the one hand, in contrast to the alpha-shaped kernel, the IF kernel never decays information with time, thereby mitigating the over-sparsity issue. On the other hand, according to the AFT mechanism, if a neuron remains inactive for an extended duration, its firing threshold will decrease, making it more susceptible to firing. The firing threshold can also rise to inhibit excessive spike generation, making the neuron maintain a stable status. Consequently, our method not only addresses the over-sparsity problem but also regulates the firing rate of deep SNNs. In summary, the proposed STD-ED algorithm effectively tackles the challenges involved in current event-driven learning algorithms. It utilizes spike timing to convey gradient information and enables the training of SNNs in a fully event-driven fashion.

% In summary, the STD-ED approach is an event-driven algorithm that leverages spike timing to carry gradient information. It effectively tackles two existing challenges by employing hardware-friendly AFT-IF neurons, with the AFT mechanism resolving the issue of over-sparsity and the IF kernel addressing the issue of gradient reversal. The integration of AFT-IF neurons into STD-ED guarantees the successful training of STD-ED in a fully event-driven manner.

\subsection{Membrane-potential-dependent event-driven learning algorithm}

\subsubsection{Neuron model} The MPD-ED learning algorithm incorporates the AFT mechanism into the widely employed Leaky-Integrate-and-Fire (LIF) model, denoted as the AFT-LIF. The membrane potential of an AFT-LIF neuron can be described in the following equations
\begin{equation}
u_i^l[t]=\tau u_i^l[t-1]\left(1-s_i^l[t-1]\right)+\sum_j \omega_{ij}^ls^{l-1}_j[t],
\label{AFTLIF}
\end{equation}
\begin{equation}
\vartheta_i^l[t]= \vartheta_0 s_i^l[t-1]+\left(\vartheta_i^l[t-1]-\alpha\right) \left(1-s_i^l[t-1]\right),
\label{d:AFT}
\end{equation}
\begin{equation}
s_i^l[t]=\mathcal{H}\left(u_i^l[t]-\vartheta_i^l[t]\right),
\label{Heaviside}
\end{equation}
where $\tau$ is the leaky factor of the membrane potential, $\alpha$ is the decay parameter of the AFT mechanism, and $\mathcal{H}$ is the Heaviside function. According to the AFT mechanism described in Eq.~\ref{d:AFT}, if an AFT-LIF neuron $i$ fails to generate a spike at time \(t-1\), its threshold undergoes a decay by $\alpha$ as described in Eq.~\ref{AFT}. Conversely, if neuron $i$ successfully generates a spike at \(t-1\), its threshold is reset to the initial value $\vartheta_0$.

\subsubsection{MPD-ED learning rule}

To train SNNs with the MPD-ED algorithm, we still need to compute the derivative of the loss function with respect to input spikes and synaptic weights, i.e., ${\partial{\mathcal L}}/{\partial s_j^{l\!-\!1}[t]}$ and ${\partial{\mathcal L}}/ {\partial\omega_{ij}^{l}}$. Further details about how the MPD-ED calculates these two terms are provided below.

\begin{figure}
\centering
\hspace{-0.04\linewidth}
\subfigure[]{
\includegraphics[scale=0.54]{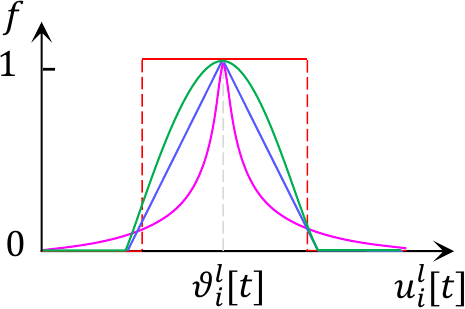}}
\subfigure[]{
\includegraphics[scale=0.54]{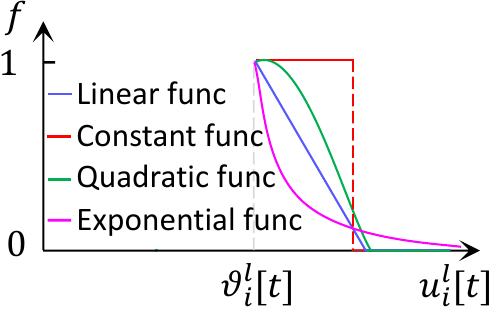}}
\caption{(a) Four typical surrogate gradient functions in SG algorithms. (b) Corresponding masked surrogate gradient functions in the MPD-ED.}
\label{fig:MPD-ED}
\end{figure}

We begin by addressing ${\partial{\mathcal L}}/{\partial s_j^{l\!-\!1}[t]}$. The calculation of this derivative can be decomposed into two components: inter-neuron dependency and intra-neuron dependency. For the inter-neuron dependency, the input spike $s_j^{l\!-\!1}[t]$ influences the loss function ${\mathcal L}$ by affecting the membrane potential and further the output spike of neuron $i$. For the intra-neuron dependency, the spike $s_j^{l\!-\!1}[t]$ influences ${\mathcal L}$ by affecting the membrane potential and further the spike activity of neuron $j$ at time $t+1$. Therefore, the derivative of the loss function with respect to the input spike can be expressed as
\begin{equation}
\begin{split}
\frac{\partial \mathcal{L}}{\partial s_j^{l\!-\!1}[t]}=&~\sum_{i}\frac{\partial \mathcal{L}}{\partial s_i^{l}[t]}~ \frac{\partial s_i^{l}[t]}{\partial u_i^{l}[t]}~\frac{\partial u_i^{l}[t]}{\partial s_j^{l\!-\!1}[t]}\\&\!+\!\frac{\partial \mathcal{L}}{\partial s_j^{l\!-\!1}[t\!+\!1]} \frac{\partial s_j^{l\!-\!1}[t\!+\!1]}{\partial u_j^{l\!-\!1}[t\!+\!1]}\frac{\partial u_j^{l\!-\!1}[t\!+\!1]}{\partial s_j^{l\!-\!1}[t]}.
\label{M:l/s}
\end{split}
\end{equation}
In both of these two dependencies, the first item can be recursively computed, and the third item can be easily obtained based on Eq.~\ref{AFTLIF}. However, due to the non-differentiability, calculating the second item in both dependencies proves to be challenging. In the domain of SG learning, this challenge is addressed by replacing it with a surrogate gradient. However, this method necessitates gradient updating at each time step regardless of spike activity, demanding substantial training resources. To mitigate this issue, we introduce the masked surrogate gradient function (MSG) to train SNNs in an event-driven manner. The proposed MSG function can be described as
\begin{equation}
\frac{\partial s_i^{l}[t]}{\partial u_i^{l}[t]}=\left\{
\begin{array}{ll}
f\!\left(u_i^l[t],\vartheta_i^l[t]\right), & {u_i^l[t]\geq \vartheta_i^l[t]},\\
0, & \text{otherwise}.
\end{array} \right.
\label{masksg}
\end{equation} 
As shown in Fig.~\ref{fig:MPD-ED}, the exact definition of the MSG function can be various, such as the constant function, linear function, exponential function, etc. Unlike conventional SG learning that performs backpropagation at each time step, the proposed MSG method conducts gradient backpropagation only when the membrane potential crosses the firing threshold. By propagating the gradient only through the generated spike, the learning cost can be significantly reduced. However, it still suffers from the problem of over-sparsity spikes. Fortunately, the proposed AFT mechanism dynamically adjusts the threshold to address the issue of over-sparsity, enabling successful training of SNN using the efficient MSG function. Finally, combined with Eqs.[\ref{M:l/s}-\ref{masksg}], the derivative of the loss function to the input spike is computationally feasible.
% i.e., $u_i^l[t]\!\geq\!\vartheta_i^l[t]$

We now compute ${\partial{\mathcal L}}/{\partial\omega_{ij}^{l}}$. The synaptic weight $\omega_{ij}^{l}$ influences the loss function ${\mathcal L}$ by affecting the membrane potential of neuron $i$, and this influence takes place at each time step, so the derivative can be described as
\begin{equation}
\frac{\partial \mathcal{L}}{\partial\omega_{ij}^l}=\sum_{t=1}^T\frac{\partial \mathcal{L}}{\partial u_i^l[t]}\frac{\partial u_i^l[t]}{\partial\omega_{ij}^l}.
\label{M:l/w}
\end{equation}
In this equation, the second item can be easily obtained following Eq.~\ref{AFTLIF}. The computation of the first item is similar to Eq.~\ref{M:l/s} that can be decomposed as
\begin{equation}
\begin{split}
\frac{\partial \mathcal{L}}{\partial u_i^{l}[t]}\!=\!\frac{\partial \mathcal{L}}{\partial s_i^{l}[t]} \frac{\partial s_i^{l}[t]}{\partial u_i^{l}[t]} \!+\!\frac{\partial \mathcal{L}}{\partial s_i^{l}[t\!+\!1]} \frac{\partial s_i^{l}[t\!+\!1]}{\partial u_i^{l}[t\!+\!1]}\frac{\partial u_i^{l}[t\!+\!1]}{\partial u_i^{l}[t]}.
\end{split}
\label{M:l/u}
\end{equation}
These two components represent inter-neuron dependency and intra-neuron dependency, respectively. In this equation, the derivative of ${\partial u_i^{l}[t\!+\!1]}/{\partial u_i^{l}[t]}$ can also be obtained following Eq.~\ref{AFTLIF} and the calculation of other items have been elucidated before. Consequently, the synaptic weight can be updated. In summary, both Eq.~\ref{M:l/s} and Eq.~\ref{M:l/w} formulate the MPD-ED learning rules, enabling the SNN to be trained successfully with the membrane potential in an event-driven manner.

\subsubsection{Analysis and summary}We thoroughly analyze how the MPD-ED resolves the challenges of gradient reversal and over-sparsity. As analyzed in Section \ref{gradient_reversal}, the gradient reversal problem arises from the computation of \(d K(\cdot)/d t\) during gradient backpropagation, where the negative value of this derivative can reverse the gradient of spike timing. Fortunately, according to Eqs.[\ref{M:l/s}-\ref{M:l/u}], the MPD-ED learning algorithm avoids the need for computing this derivative. Consequently, the MPD-ED method inherently avoids the gradient reversal problem. Despite the circumventions of the gradient reversal problem, the MPD-ED still suffers from the over-sparsity issue. The MPD-ED method tackles this issue by incorporating the AFT mechanism into the widely used LIF neuron model. This incorporation allows spiking neurons to adaptively adjust the firing threshold, not only addressing the over-sparsity problem but also preventing excessive spike generation. Overall, the MPD-ED is the first event-driven algorithm that utilizes the membrane potential as the learning signal, which achieves sparse event-driven backpropagation through the proposed MSG function and effectively addresses over-sparsity by incorporating the AFT mechanism.

\section{Experiments}

In this section, we begin by presenting the experiment setup and implementation details. Subsequently, we evaluate the performance of our methods by comparing them to existing methods on multiple benchmark datasets. Finally, we conduct ablation studies to verify the effectiveness of the crucial components within the two proposed algorithms.

\subsection{Experiment setup}
\label{setup}

\subsubsection{Dataset} 
We investigate the efficacy of our methods on both static datasets, including F-MNIST~\cite{xiao2017fashion}, CIFAR-10~\cite{krizhevsky2009learning}, and CIFAR-100~\cite{krizhevsky2009learning}, as well as neuromorphic datasets such as N-MNIST~\cite{orchard2015converting}, DVS-Gesture~\cite{amir2017low}, and DVS-CIFAR10~\cite{li2017cifar10}. These datasets have been extensively employed in the machine learning and neuromorphic computing communities as standard benchmarks for evaluating various learning algorithms. Before introducing the experiments, we provide a concise overview of each dataset. The static F-MNIST dataset comprises 70K grayscale images, with a division of 60K for training and 10K for testing. Each grayscale image has a spatial resolution of 28$\times$28. The CIFAR dataset is a more complex static dataset, which provides 50K training images and 10K testing images. Each image has a resolution of 32$\times$32. The N-MNIST dataset is an event-based version of the MNIST dataset, which comprises 60K training samples and 10K testing samples. Each sample consists of two channels and features a resolution of 34$\times$34. DVS-Gesture and DVS-CIFAR10 are both neuromorphic datasets captured by the DVS camera, which have the same spatial resolution of 128$\times$128. The DVS-Gesture dataset contains 1,464 samples, of which 1,176 are allocated for training and 288 for testing. The DVS-CIFAR10 dataset is currently the most challenging neuromorphic dataset, which provides 9K training samples and 1K testing samples. When preprocessing the DVS-CIFAR10, we apply data augmentation techniques as proposed in~\cite{li2022neuromorphic}.

\subsubsection{Network architecture} In the following, we describe the architectures employed for each dataset. For the static F-MNIST dataset, we adopt the architecture of 32C5-P2-64C5-P2-1024, following previous works~\cite{zhang2020temporal,zhu2022training} for the purpose of comparison. The numbers followed by `C' and `P' represent the kernel size of the convolution filter and pooling filter, respectively. For the static CIFAR-10 and CIFAR-100 datasets, we investigate several classical architectures such as VGGNet~\cite{simonyan2014very}, SEW-ResNet~\cite{fang2021deep}, and MS-ResNet~\cite{hu2021advancing}. In the STD-ED algorithm, we employ VGG11 and SEW-ResNet-14. In the MPD-ED algorithm, we employ VGG11(w/o. FC) where unnecessary fully connected (FC) layers are removed based on the original VGG11, as well as MS-ResNet-18. For the N-MNIST dataset, we employ the architecture of 32C5-P2-64C5-P2-1024. For the DVS-Gesture and DVS-CIFAR10 datasets, we implement the architecture of VGG11$^\star$. As for the pooling layer in the above architectures, we employ the adjusted average pooling in the STD-ED~\cite{zhu2022training,zhu2023exploring} and the max pooling in the MPD-ED.

\subsubsection{Implementation details} We implement the STD-ED in discrete time steps to leverage available deep learning frameworks~\cite{zhu2022training,zhu2023exploring}. Specifically, binary spikes are used for feedforward computations, while spike times are used for gradient backpropagation. The input image is encoded using the direct coding scheme~\cite{deng2022temporal,zhu2022training}, where spike currents are utilized to represent pixel intensities. To guide the training process of the STD-ED, we employ an enhanced spike count loss, which measures the discrepancy between actual and desired spike numbers of the output~\cite{zhu2023exploring}. To apply this loss function, we need to specify desired spike numbers for target(non-target) neurons. For the datasets mentioned in Table ~\ref{tab: results}, these desired spike numbers are set to 5(1), 10(1), 15(1), 15(2), 15(2), and 15(2), respectively. Moreover, we set the initial threshold $\vartheta_0$ and the decay term $\alpha$ in the AFT mechanism to 1 and $1/T$, where $T$ represents the time step and we present it in Table~\ref{tab: results}. In addition, the training process for the CIFAR dataset adopts the SGD optimizer, while the other datasets employ the AdamW optimizer. During training, we employ a cosine annealing learning rate curve and set the batch size to 50 for all datasets.

In the MPD-ED algorithm, we follow the direct coding scheme, the AFT setup, and the cosine annealing learning rate curve as utilized in the STD-ED. The MPD-ED algorithm incorporates the temporal efficient training (TET) loss function, which constrains the output of the network at each time step to closely match the target distribution~\cite{deng2022temporal}. The setting of the hyperparameter in the TET loss follows the official specification. In addition, the MPD-ED method needs to specify the leaky factor for the AFT-LIF model, which is set to 0.5 in our implementation. During the training process, we employ the AdamW optimizer for all datasets and set the batch size to 512 for static image datasets as well as 64 for neuromorphic datasets.

\begin{table*}
\centering
\caption{Classification performance comparison on static image datasets and neuromorphic datasets.}
\renewcommand{\arraystretch}{1.3}
\setlength{\tabcolsep}{3mm}
\begin{threeparttable}
\begin{tabular}{l|lllll}
\hline
 & Method & Network Architecture & Event Driven & Time Steps & Accuracy \\ \hline
 & Kheradpisheh et al.~\cite{kheradpisheh2020temporal} & 784-1000 &\Checkmark& 256 & 88.00\% \\
 & Kheradpisheh et al.~\cite{kheradpisheh2022bs4nn} & 784-1000 &\Checkmark& 256 & 87.30\% \\
 & Zhang et al.~\cite{zhang2021rectified} & 16C5-P2-32C5-P2-800-128 &\Checkmark& 450 & 90.10\% \\
 & Zhang et al.~\cite{zhang2020temporal} & 32C5-P2-64C5-P2-1024 &\Checkmark& 5 & 92.83\% \\
 & Zhu et al.~\cite{zhu2022training} & 32C5-P2-64C5-P2-1024 &\Checkmark& 5 & 93.28\% \\
 & Zhu et al.~\cite{zhu2023exploring} & 32C5-P2-64C5-P2-1024 &\Checkmark& 5 & 94.03\% \\
\cline{2-6} 
 & \textbf{This work(STD-ED)} & {32C5-P2-64C5-P2-1024} &\Checkmark& {5} & \textbf{94.05\%} \\
\multirow{-8}{*}{F-MNIST} & \textbf{This work(MPD-ED)} & {32C5-P2-64C5-P2-1024} 
 &\Checkmark& {5} & \textbf{94.04\%} \\ \hline 
 & Deng et al.~\cite{deng2022temporal} &ResNet-19 &\XSolidBrush& 6 &94.50\% \\
 & Hu et al.~\cite{hu2021advancing}$^\ddagger$ &MS-ResNet-18&\XSolidBrush& 6 & 94.92\% \\
 & Xiao et al.~\cite{xiao2022online} &VGG11$^\star$&\XSolidBrush& 6 & 93.73\% \\
 & Meng et al.~\cite{meng2023towards} &ResNet-18&\XSolidBrush& 6 & 94.59\% \\ \cline{2-6} 
 & Park et al.~\cite{park2021training} & VGG16 &\Checkmark& 544 & 91.90\% \\
 & Wei et al.~\cite{wei2023temporal} & VGG16 &\Checkmark& 160 & 93.05\% \\
 & Zhang et al.~\cite{zhang2020temporal} & CIFARNet &\Checkmark& 5 & 91.41\% \\
 & Zhu et al.~\cite{zhu2022training} & VGG11 &\Checkmark& 12 & 92.10\% \\
 & Zhu et al.~\cite{zhu2022training} & SEW-ResNet-14 &\Checkmark& 12 & 92.45\% \\ 
 & Zhu et al.~\cite{zhu2023exploring} & VGG11 &\Checkmark& 12 & 93.54\% \\ 
 \cline{2-6} 
 & {} & {VGG11} &\Checkmark& {12} & \textbf{94.33\%} \\
 & \multirow{-2}{*}{\textbf{This work(STD-ED)}} & {SEW-ResNet-14} &\Checkmark& {12} & \textbf{93.85\%} \\ 
 & {} & {VGG11(w/o.FC)} &\Checkmark& {5} & \textbf{94.51\%} \\
\multirow{-14}{*}{CIFAR-10} & \multirow{-2}{*}{\textbf{This work(MPD-ED)}} & {MS-ResNet-18} &\Checkmark& {6} & \textbf{94.84\%} \\ \hline 
 & Deng et al.~\cite{deng2022temporal} &ResNet-19 &\XSolidBrush& 6 & 74.72\% \\
 & Hu et al.~\cite{hu2021advancing}$^\ddagger$ &MS-ResNet-18 &\XSolidBrush& 6 & 76.41\% \\
 & Xiao et al.~\cite{xiao2022online} &VGG11$^\star$ &\XSolidBrush& 6 & 71.11\% \\
 & Meng et al.~\cite{meng2023towards} &ResNet-18 &\XSolidBrush& 6 & 74.67\%
 \\ \cline{2-6} 
 & Park et al.~\cite{park2021training} &VGG16 &\Checkmark& 544 & 65.98\% \\
 & Wei et al.~\cite{wei2023temporal} &VGG16 &\Checkmark& 160 & 69.66\% \\
 & Zhu et al.~\cite{zhu2022training} &VGG11 &\Checkmark& 16 & 63.97\% \\
 & Zhu et al.~\cite{zhu2023exploring} &VGG11 &\Checkmark& 16 & 70.50\% \\ 
 \cline{2-6} 
 & {} & {VGG11} &\Checkmark& {16} & \textbf{73.01\%} \\
 & \multirow{-2}{*}{\textbf{This work(STD-ED)}} & {SEW-ResNet-14} &\Checkmark& {16} & \textbf{71.63\%} \\ 
 & {} & {VGG11(w/o.FC)} &\Checkmark& {5} & \textbf{75.33\%} \\
\multirow{-12}{*}{CIFAR-100} & \multirow{-2}{*}{\textbf{This work(MPD-ED)}} 
 & {MS-ResNet-18} &\Checkmark& {6} & \textbf{77.29\%} \\ \hline 
 \multirow{5}{*}{N-MNIST} 
 & Zhang et al.~\cite{zhang2020temporal} & 12C5-P2-64C5-P2 &\Checkmark& 100 & 99.40\% \\
 & Zhu et al.~\cite{zhu2022training} & 12C5-P2-64C5-P2 &\Checkmark& 30 & 99.39\% \\
 & Zhu et al.~\cite{zhu2023exploring} & 12C5-P2-64C5-P2 &\Checkmark& 30 & 99.39\% \\
 \cline{2-6} 
 & \textbf{This work(STD-ED)} & {12C5-P2-64C5-P2} &\Checkmark& {30} & \textbf{99.40\%} \\
 & \textbf{This work(MPD-ED)} & {12C5-P2-64C5-P2} &\Checkmark& {10} & \textbf{99.36\%} \\ \hline 
 & Xiao et al.~\cite{xiao2022online} & VGG11$^\star$&\XSolidBrush &20  & 96.88\% \\
 & Meng et al.~\cite{meng2023towards} & VGG11$^\star$&\XSolidBrush &20  & 98.62\% \\
 \cline{2-6} 
 & Zhu et al.~\cite{zhu2023exploring} & VGG11&\Checkmark & 20 & 97.22\% \\
 \cline{2-6} 
 & \textbf{This work(STD-ED)} &  VGG11$^\star$ &\Checkmark& {20} & \textbf{98.96\%} \\
 \multirow{-5}{*}{DVS-Gesture} & \textbf{This work(MPD-ED)} & {VGG11$^\star$} &\Checkmark& {16} & \textbf{97.92\%} \\
 \hline 
 & Guo et al.~\cite{guo2023membrane} &ResNet-20 &\XSolidBrush & 10 & 78.80\% \\
 & Deng et al.~\cite{deng2022temporal} &VGG11$^\star$ &\XSolidBrush &10 & 83.32\% \\
 & Xiao et al.~\cite{xiao2022online} &VGG11$^\star$ &\XSolidBrush & 10  & 76.30\% \\
 & Wang et al.~\cite{wang2023ssf} &VGG11$^\star$ &\XSolidBrush & 20 & 78.00\% \\
 \cline{2-6} 
 & Zhu et al.~\cite{zhu2023exploring} & VGG11 &\Checkmark &20  & 76.30\% \\
 \cline{2-6} 
 & \textbf{This work(STD-ED)} & {VGG11$^\star$} &\Checkmark& {20} & \textbf{77.30\%} \\
 % & \textbf{This work(STD-ED)} & {VGG11(w/o.FC)} &\Checkmark& {16} & \textbf{\textcolor{red}{75.40\%}} \\
 \multirow{-5}{*}{DVS-CIFAR10} & \textbf{This work(MPD-ED)} & {VGG11$^\star$} &\Checkmark& {10} & \textbf{81.50\%}
 % \multirow{-5}{*}{DVS-CIFAR10} & \textbf{This work(MPD-ED)} & {VGG11(w/o.FC)} &\Checkmark& {10} & \textbf{81.10\%}
 \\ \hline
\end{tabular}
\begin{tablenotes}
\item[]$^\ddagger$: Self-implementation results with open-source code.
\item[]CIFARNet: 128C3-256C3-P2-512C3-P2-1024C3-512C3-1024-512.
\item[]VGG11$^\star$: 64C3-128C3-P2-256C3-256C3-P2-512C3-512C3-P2-512C3-512C3-P2.
\item[]VGG11: 128C3-128C3-P2-256C3-256C3-256C3-P2-512C3-512C3-512C3-P2-2048-2048.
\end{tablenotes}
\end{threeparttable}
\label{tab: results}
\end{table*}

\subsection{Performance comparison}

In order to thoroughly evaluate the effectiveness of our approaches, we perform a comprehensive benchmark of our approaches against existing learning algorithms, including SG methods and event-driven methods. In the following, we provide a detailed analysis of these two comparisons.

We first compare our work with widely used SG methods on several datasets, including CIFAR-10, CIFAR-100, DVS-Gesture, and DVS-CIFAR10. On the CIFAR-10 dataset, Hu et al.~\cite{hu2021advancing} achieve a state-of-the-art (SOTA) accuracy of 94.92\% with MS-ResNet-18. In our methods, the MPD-ED achieves an accuracy of 94.84\% with the same structure, and the STD-ED achieves an accuracy of 94.33\% with VGG11. Although our methods may not be SOTA, they achieve comparable performance to SG methods while significantly reducing training costs due to their efficient event-driven nature. On the CIFAR-100 dataset, Hu et al.~\cite{hu2021advancing} achieve the previously best accuracy of 76.41\% with MS-ResNet-18. In our methods, the MPD-ED achieves a SOTA accuracy of 77.29\% with the same structure, and the STD-ED achieves an accuracy of 73.01\% with VGG11. Even with sparse event-driven propagation, the MPD-ED exhibits superior performance to SG methods. On the DVS-Gesture dataset, Meng et al.~\cite{meng2023towards} report a previously best accuracy of 98.62\% with VGG11$^\star$. Under the same structure, the MPD-ED achieves an accuracy of 97.92\%, and the STD-ED achieves a SOTA accuracy of 98.96\%. On the DVS-CIFAR10 dataset, Deng et al.~\cite{deng2022temporal} report a SOTA accuracy of 83.32\% with VGG11$^\star$. Under the same structure, the MPD-ED achieves an accuracy of 81.50\%, and the STD-ED achieves an accuracy of 77.3\%. Although there may be some gaps between our work and Deng et al.~\cite{deng2022temporal} on DVS-CIFAR10, we have still achieved satisfactory results with extremely low training cost. In conclusion, our two methods have demonstrated comparable or even superior performance to well-established SG methods while requiring lower training costs. This outcome holds significant meaning in the field of event-driven algorithms for SNNs.

We now compare our work with existing event-driven methods, focusing primarily on the research by Zhu et al.~\cite{zhu2023exploring}, as it represents the state-of-the-art in the field of event-driven learning. On the F-MNIST dataset, Zhu et al.~\cite{zhu2023exploring} achieve an accuracy of 94.03\%. Under the same structure, the MPD-ED achieves an accuracy of 94.04\%, and the STD-ED exhibits an accuracy of 94.05\%. Both of these results surpass the performance achieved by Zhu et al.\cite{zhu2023exploring}. The performance gap is not conspicuous on the simple dataset, but it further widens when applied to complex datasets. On the CIFAR-10 dataset, Zhu et al.~\cite{zhu2023exploring} achieve an accuracy of 93.54\% with VGG11. In our methods, the MPD-ED achieves an accuracy of 94.84\% with MS-ResNet-18, and the STD-ED achieves an accuracy of 94.33\% with the same VGG11. This demonstrates a performance improvement of 1.3\% for the MPD-ED and 0.79\% for the STD-ED. On the CIFAR-100 dataset, Zhu et al.~\cite{zhu2023exploring} achieve an accuracy of 70.50\% with VGG11. In our methods, the MPD-ED achieves an accuracy of 77.29\% with MS-ResNet-18, and the STD-ED achieves an accuracy of 73.01\% with the same VGG11. This indicates a significant improvement in terms of accuracy, with 6.79\% for the MPD-ED and 2.51\% for the STD-ED. On simple neuromorphic dataset N-MNIST, Zhu et al.~\cite{zhu2023exploring} achieve an accuracy of 99.39\%. Under the same structure, the MPD-ED achieves an accuracy of 99.36\% with fewer time steps, and the STD-ED achieves a SOTA accuracy of 99.40\%. On complex neuromorphic datasets, Zhu et al.~\cite{zhu2023exploring} is the only event-driven work that reports performance, achieving accuracies of 97.22\% on DVS-Gesture and 76.30\% on DVS-CIFAR10. In our methods, the MPD-ED achieves accuracies of 97.92\% and 81.50\% on these datasets, outperforming Zhu et al.~\cite{zhu2023exploring} by 0.7\% and 5.2\% using shallower architecture and fewer time steps. The STD-ED achieves accuracies of 98.96\% and 77.3\%, surpassing Zhu et al.~\cite{zhu2023exploring} by 1.74\% and 1\%, respectively. In conclusion, our methods achieve SOTA results among the existing event-driven algorithms, significantly raising the performance of event-driven algorithms to a new level.

The emergence of STD-ED and MPD-ED is of great significance in the field of learning algorithms for SNNs. In comparison to well-established SG methods, our approaches stand out for their energy efficiency, as gradient backpropagation is performed only when there are spike emissions. This energy efficiency significantly reduces resource demands during the training process, leading to lower memory usage and power consumption. Moreover, when compared to event-driven methods, our work demonstrates substantial performance improvement, yielding comparable or even superior performance to SG methods. This performance improvement propels the advancement of event-driven algorithms, paving the way for the development of energy-saving learning algorithms and neuromorphic hardware.

\subsection{Ablation study}

To prove the effectiveness of our methods, we conduct ablation experiments on essential components within the STD-ED and the MPD-ED. Ablation experiments are performed on the CIFAR-10 and CIFAR-100 datasets using the VGGNet, and the experimental setup follows the description provided in Section \ref{setup}.

\begin{figure*}[htbp]
\centering
\subfigure[]{
\includegraphics[width=4.2cm,height=3.7cm]{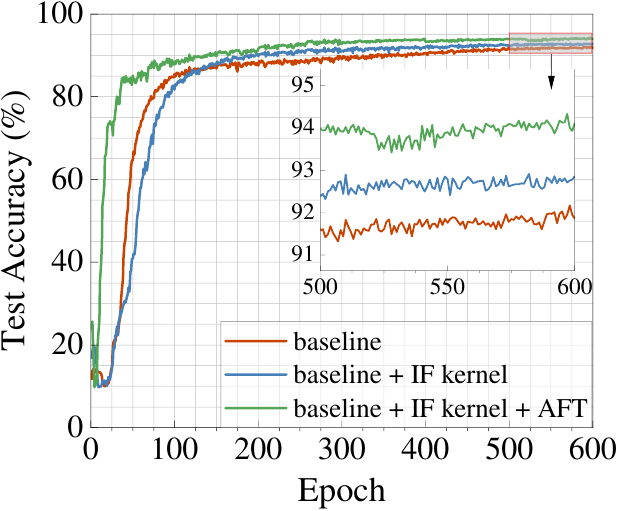}}
\subfigure[]{
\includegraphics[width=4.1cm,height=3.65cm]{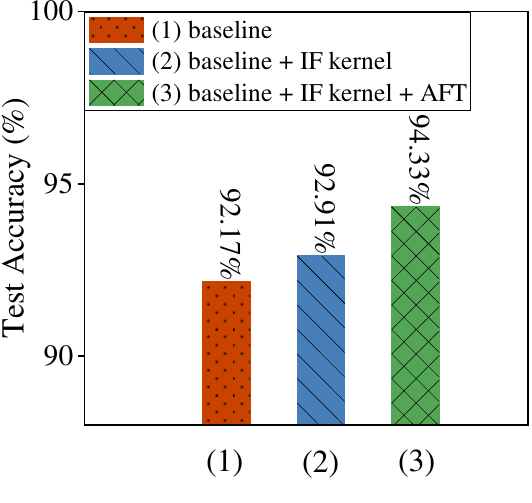}}
\subfigure[]{
\includegraphics[width=4.2cm,height=3.7cm]{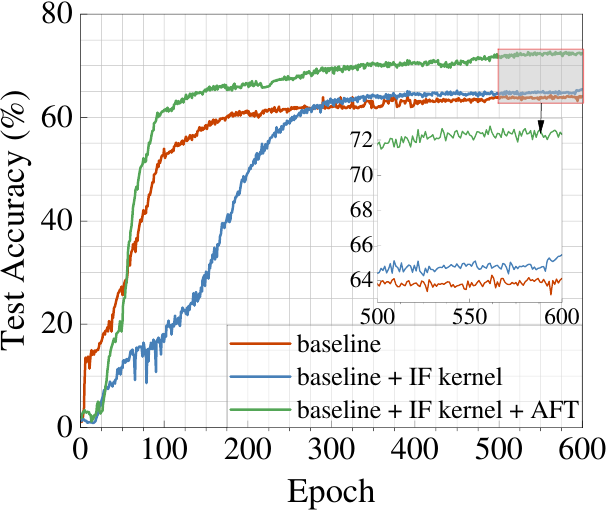}}
\subfigure[]{
\includegraphics[width=4cm,height=3.65cm]{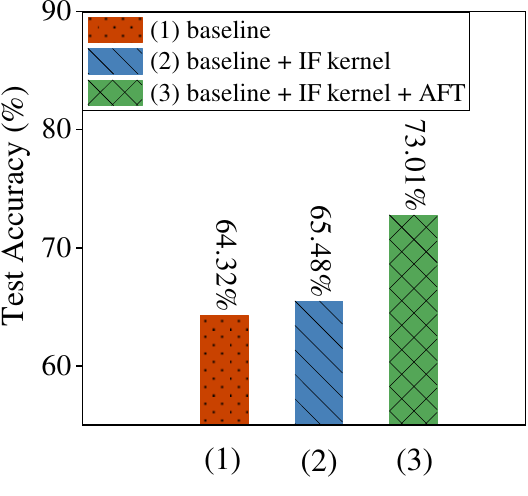}}
\caption{Ablation studies of the STD-ED on CIFAR datasets, where the IF kernel and the AFT mechanism are ablated. (a) Convergence curves of three comparative methods on CIFAR-10. (b) Accuracy of three comparative methods on CIFAR-10. (c) Convergence curves of three comparative methods on CIFAR-100. (d) Accuracy of three comparative methods on CIFAR-100.}
\label{ablate:TED}
\end{figure*}
\begin{figure*}
\centering
\subfigure[]{
\includegraphics[width=4.12cm,height=3.68cm]{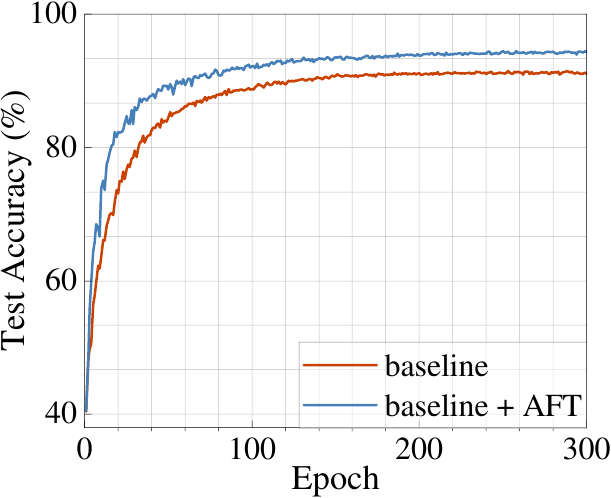}}
% \hspace{0.4mm}
\subfigure[]{
\includegraphics[width=4cm,height=3.6cm]{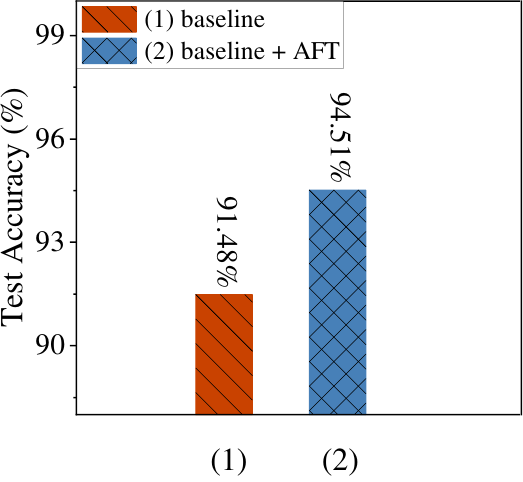}}
\hspace{0.6mm}
\subfigure[]{ 
\includegraphics[width=4.25cm,height=3.7cm]{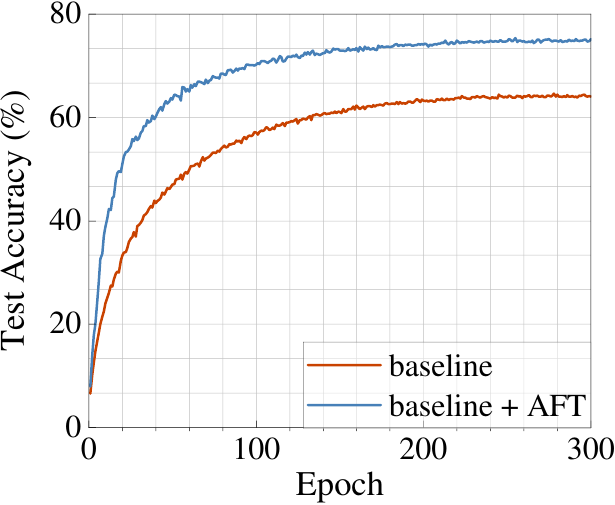}}
\subfigure[]{
\includegraphics[width=3.9cm,height=3.65cm]{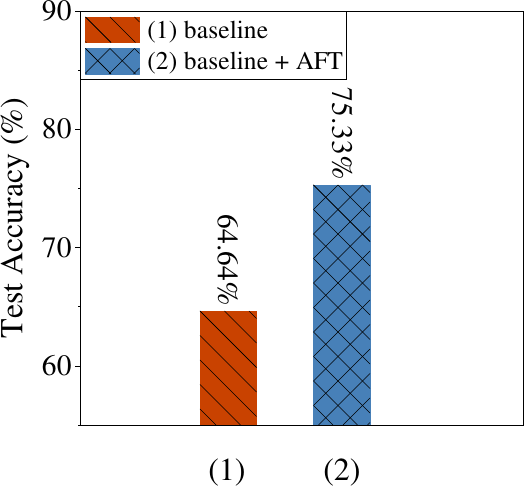}}
\caption{Ablation studies of the MPD-ED on CIFAR datasets, where only the AFT mechanism is ablated. (a) Convergence curves of two comparative methods on CIFAR-10. (b) Accuracy of two comparative methods on CIFAR-10. (c) Convergence curves of two comparative methods on CIFAR-100. (d) Accuracy of two comparative models on CIFAR-100.}
\label{ablate:AED}
\end{figure*}

\subsubsection{Ablation of the STD-ED}The STD-ED addresses the challenges of over-sparsity and gradient reversal by utilizing the IF kernel and the AFT mechanism, respectively. Therefore, we ablate two components within the STD-ED: the IF kernel and the AFT mechanism. We choose the baseline method employing the alpha-shaped kernel and the fixed threshold for comparison, and it resolves the gradient reversal issue by using an exponential function in the backward process~\cite{zhu2022training,zhu2023exploring}. Consequently, we compare three methods: baseline, baseline replaced with the IF kernel, and baseline replaced with the IF kernel and the AFT (namely the STD-ED).

During the learning process of the STD-ED, we record and plot two metrics for comparative analysis: convergence curve and accuracy. On the CIFAR-10 dataset, as shown in Fig.~\ref{ablate:TED}, the STD-ED method achieves the fastest convergence and the top-1 accuracy of 94.33\%. Moreover, the baseline method demonstrates the second fastest convergence speed. However, despite its relatively fast convergence, the baseline method exhibits the poorest accuracy of 92.17\% among the three methods. In contrast, the baseline replaced with the IF kernel displays better potential in the whole learning process than the baseline, eventually achieving the top-2 accuracy of 92.91\%. The same phenomenon can be observed in the CIFAR-100 dataset. As a result, these ablation experiments prove the effectiveness of the IF kernel and the AFT mechanism within the STD-ED. On the one hand, it affirms that the IF kernel is more suitable than the alpha-shaped kernel for the spike timing-based event-driven learning of SNNs. On the other hand, the efficacy of the AFT in mitigating the over-sparsity problem has been proven since it enhances convergence speed and improves overall performance.

\subsubsection{Ablation of the MPD-ED}Among the two challenges faced by event-driven learning algorithms, the proposed MPD-ED method only encounters the over-sparsity problem and resolves it by utilizing the AFT mechanism. Therefore, we ablate only the AFT mechanism within the MPD-ED. We choose the baseline method employing the LIF model with a fixed threshold for comparison. Consequently, we perform comparative analyses between two methods: baseline, and baseline replaced with the AFT (namely the MPD-ED).

In the ablation of the MPD-ED, we also plot two metrics of convergence curve and accuracy for comparative analysis. On the CIFAR-10 dataset, as displayed in Fig.~\ref{ablate:AED}, the MPD-ED demonstrates the fastest convergence and the highest accuracy, i.e., 94.51\%. Noteworthy, the accuracy achieved by the MPD-ED is comparable to that of well-established SG learning algorithms. In contrast, the baseline method exhibits slower convergence speed and lower accuracy than the MPD-ED. The same phenomenon is no exception in the CIFAR-100 dataset. As a result, these ablation experiments reconfirm the effectiveness of the proposed AFT mechanism. The effectiveness of the AFT mechanism stems from its capability that adaptively adjust the firing threshold of spiking neurons, which helps to prevent the occurrence of the over-sparsity problem, giving the network more learning opportunities and ultimately resulting in high performance.

\begin{figure*}
\centering 
\subfigure[Forward]{\label{fig:fw}
\includegraphics[width=0.23\textwidth]{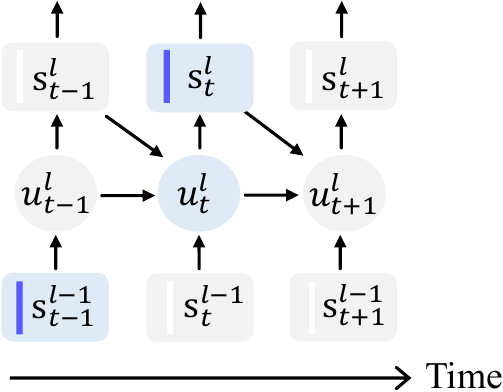}}
\subfigure[STBP backward]{
\includegraphics[width=0.23\textwidth]{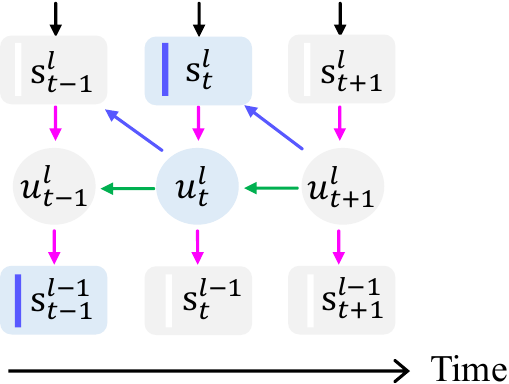}}
\subfigure[MPD-ED backward]{
\includegraphics[width=0.23\textwidth]{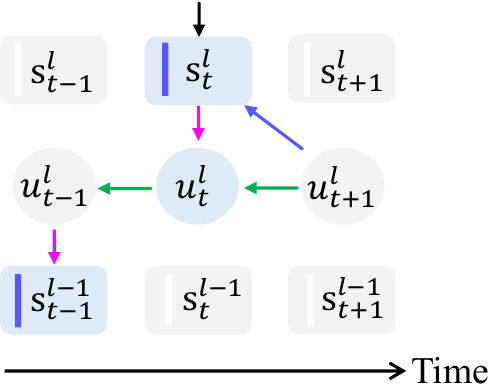}}
\subfigure[STD-ED backward]{
\includegraphics[width=0.22\textwidth]{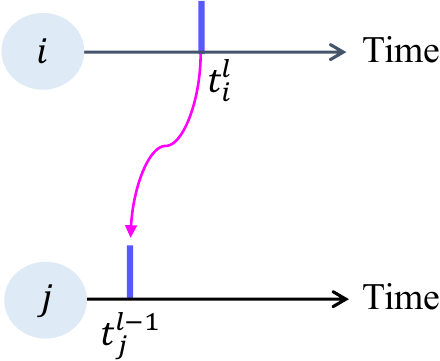}}
\caption{Computational graphs of a single FC layer, only depicting the propagation between neuron \(j\) in layer \(l-1\) and neuron \(i\) in layer \(l\). The gradient computation involves three aspects: \textit{inter-neuron BP}, \textit{intra-neuron BP}, and \textit{spike-induced BP}, represented by magenta, green, and blue arrows respectively.}
\label{fig:algcomp}
\end{figure*}

\section{Energy consumption analysis}

In this section, we explore the energy efficiency of our work through theoretical analysis and hardware deployment. In the theoretical analysis, we study the training complexity of our methods while conducting validation experiments to prove their efficiency and effectiveness. In addition, we deploy the simple and efficient MPD-ED on the neuromorphic chip to further demonstrate its efficiency and applicability.

\subsection{Theoretical analysis}

\subsubsection{Training complexity}In order to demonstrate the low-power nature of event-driven learning, we analyze the training complexity of three algorithms: SG, MPD-ED, and STD-ED. In the field of SG learning algorithms, we choose the widely used STBP method~\cite{wu2018spatio} as a representative for analysis. The gradient computation of these three algorithms in the training process involves three aspects: \textit{inter-neuron BP}, \textit{intra-neuron BP}, and \textit{spike-induced BP}. Specifically, \textit{inter-neuron BP} depicts the error signal from the next layer, \textit{intra-neuron BP} represents the propagation of intra-neuron dynamics, and \textit{spike-induced BP} describes the propagation caused by the reset mechanism. We depict computation graphs of three algorithms in Fig.~\ref{fig:algcomp}, where three types of gradient computation are indicated by magenta, green, and blue arrows, respectively. In the following, we analyze the training complexity of three algorithms in detail.

We focus on analyzing the training complexity of a single FC layer for the sake of simplicity, and this analysis can be extended to other layers as well as the entire network. Assuming that one FC layer consists of $M$ input neurons and $N$ output neurons, with $\zeta_{i}$ and $\zeta_{o}$ representing the average spike activity of input and output neurons, respectively. In the STBP learning process, as shown in Fig.~\ref{fig:algcomp}(b), there are three types of gradient computation. Firstly, \textit{inter-neuron BP} is performed at each time step, incurring a complexity of $O(MNT)$. Secondly, \textit{intra-neuron BP} is dependent on the number of time steps, giving rise to a complexity of $O(NT)$. Thirdly, \textit{spike-induced BP} also takes place at each time step, resulting in a complexity of $O(NT)$. Consequently, the overall training complexity of the STBP is $O(MNT+NT+NT)$. The MPD-ED method is an event-driven learning algorithm that performs gradient computation only upon spike emission. As depicted in Fig.~\ref{fig:algcomp}(c), the MPD-ED also involves three types of gradient computation, however, which differ from those of the STBP due to event-driven learning. Firstly, \textit{inter-neuron BP} only occurs at the time step when a spike is emitted, with each input spike participating in this propagation only once, yielding a complexity of $O(\zeta_i MNT)$. Secondly, \textit{intra-neuron BP} remains consistent with the STBP, leading to a complexity of $O(NT)$. Thirdly, \textit{spike-induced BP} also occurs upon spike emission, resulting in a complexity of $O(\zeta_oNT)$. Consequently, the overall training complexity of the MPD-ED is $O(\zeta_i MNT+NT+\zeta_oNT)$. The training process of the STD-ED, as illustrated in Fig.~\ref{fig:algcomp}(d), involves only inter-neuron BP. The STD-ED is also an event-driven learning algorithm, with its \textit{inter-neuron BP} remaining consistent with that of the MPD-ED. Consequently, the overall training complexity of the STD-ED is $O(\zeta_iMNT)$. Table~\ref{table:complexity} provides an overview of the training complexity for these three algorithms.

\begin{figure*}
\centering
\subfigure[]{\label{fig:net}
\includegraphics[scale=0.35]{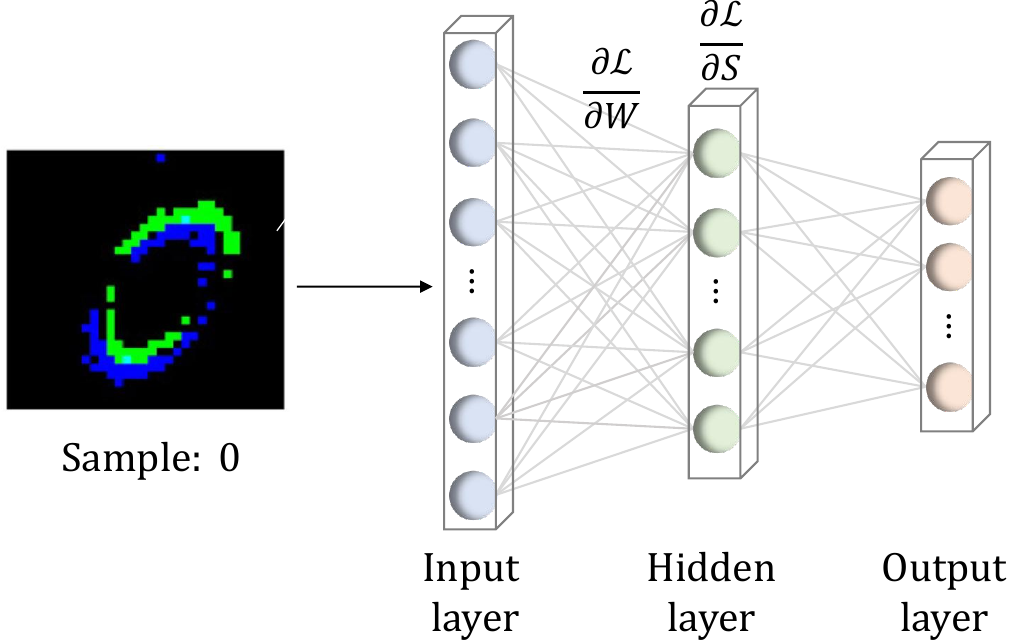}}
\hspace{0.016\linewidth}
\subfigure[]{
\includegraphics[scale=0.27]{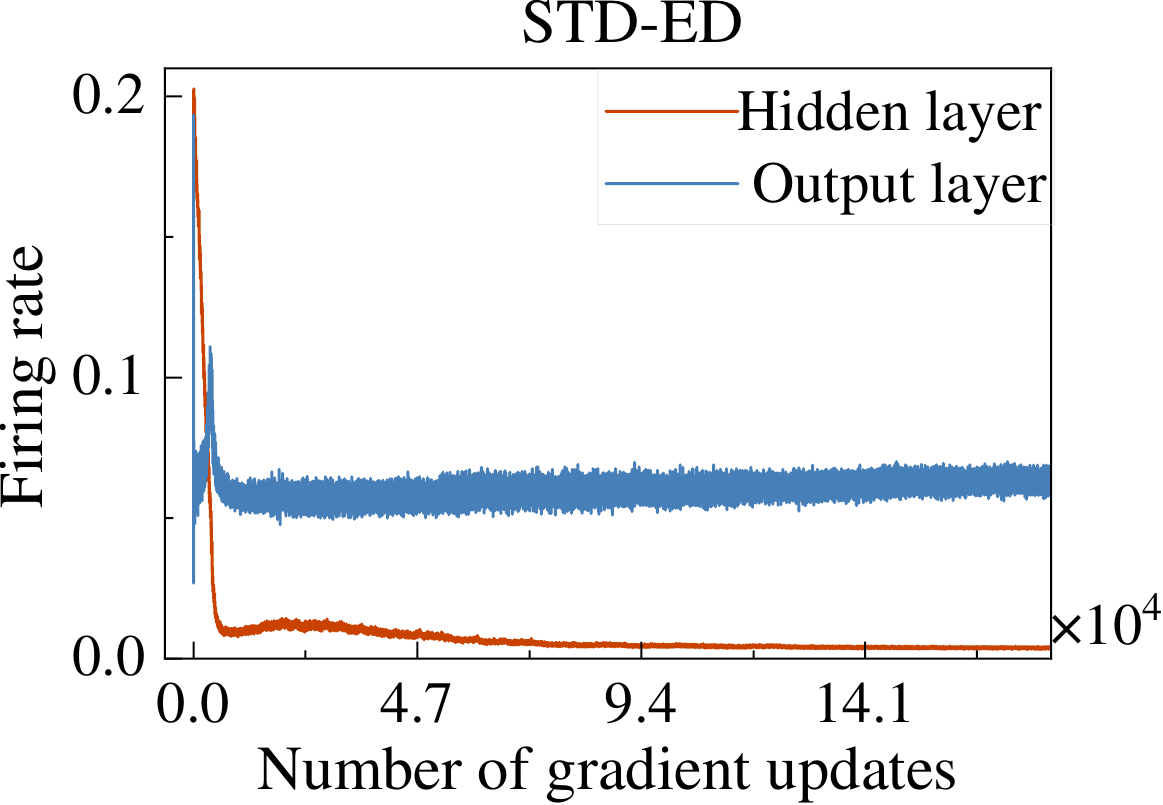}}
\subfigure[]{
\includegraphics[scale=0.27]{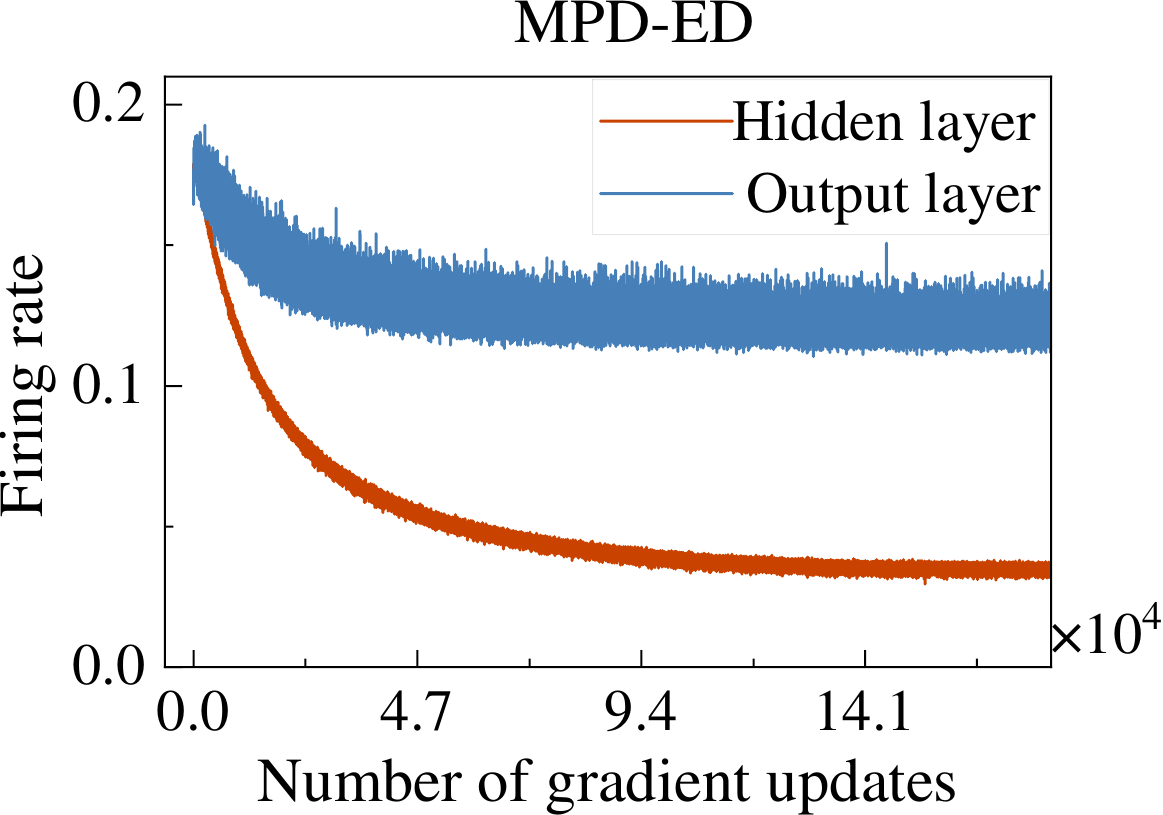}}
\subfigure[]{\label{fig:gradS}
\includegraphics[scale=0.68]{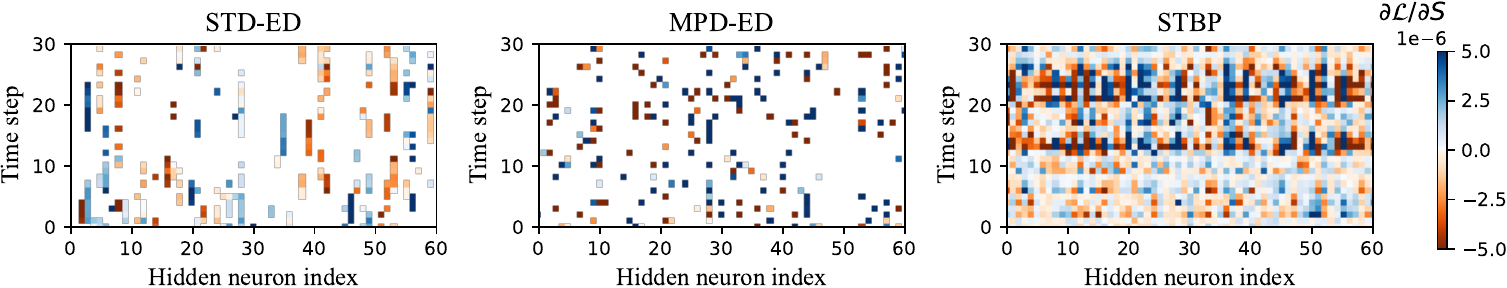}}
\subfigure[]{\label{fig:gradW}
\includegraphics[scale=0.68]{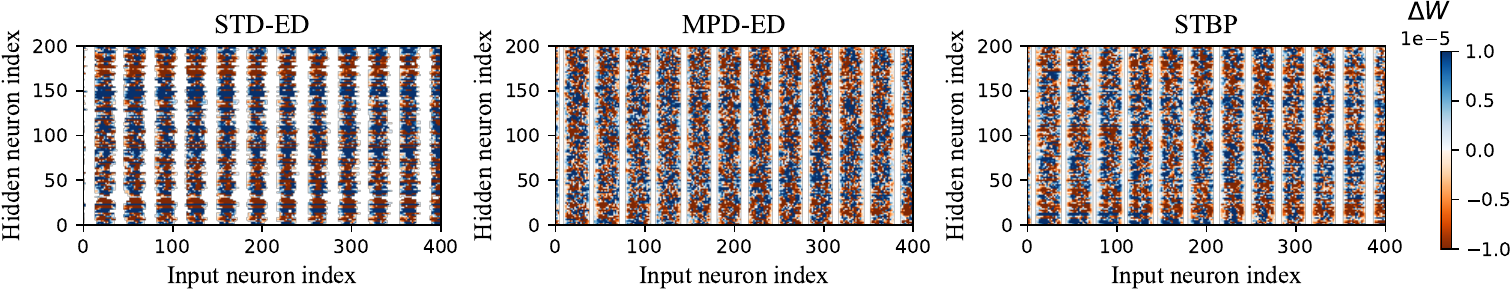}}
\caption{Validation experiment. (a) Network architecture. (b) Average spike activity in each layer during the training process for the STD-ED. (c) Average spike activity in each layer during the training process for the MPD-ED. (d) Gradient visualization of the loss function on spikes (in the hidden layer), i.e., $\partial \mathcal{L}/\partial S$. (e) Gradient visualization of the loss function on weights (between the input layer and the hidden layer), i.e., $\Delta W$. In subfigures (d,e), gradients for visualization are selected from the first sample in the first training epoch.}
\label{fig:energy} 
\end{figure*}

\subsubsection{Validation experiments} To substantiate the efficiency and effectiveness of our methods, we perform validation experiments utilizing three algorithms: STD-ED, MPD-ED, and STBP. As depicted in Fig.~\ref{fig:energy}(a), experiments are performed with the structure of 34$\times$34$\times$2-200-10 on the N-MNIST dataset, undergoing training for 150 epochs with 30 time steps. Note that the MPD-ED employs the cross-entropy loss function, with its last layer output spikes. All other experimental setups remain consistent with the settings in Section \ref{setup}.

We first assess the improved efficiency of our approaches. The training complexity in Table \ref{table:complexity} involves the calculation of average spike activity, i.e., $\zeta_i$ and $\zeta_o$, so we record it during the training process. We depict the average spike activity of each layer for the STD-ED and MPD-ED in Fig.~\ref{fig:energy}(b-c), and present the average spike activity across the whole training period in Table ~\ref{table:energy}. Leveraging both training complexity and average spike activity, we are able to quantitatively assess the reduction in training complexity achieved by our approaches. Our analysis focuses on the second FC layer (from the hidden to the output layer), where $M=200$, $N=10$, and $T=30$. Based on the recorded spike activity, for the STD-ED, $\zeta_i=0.0086$ and $\zeta_o=0.0604$; while for the MPD-ED, $\zeta_i=0.0527$ and $\zeta_o=0.1287$. As a result, the MPD-ED achieves a 94.22\% reduction in complexity compared to the STBP, while the STD-ED achieves an even higher efficiency with a 99.14\% reduction. These results indicate that our methods significantly reduce the demands on training resources, leading to lower energy consumption during the training process.

We now showcase the effective training of our approaches. Throughout the training process, we record the gradients of the loss function with respect to spikes (in the hidden layer) and weights (in the first FC layer), as labeled $\partial \mathcal{L}/\partial S$ and $\partial \mathcal{L}/\partial W$ in Fig.~\ref{fig:net}. We depict $\partial \mathcal{L}/\partial S$ in Fig.~\ref{fig:gradS}, where the STD-ED and MPD-ED propagate spike gradients only upon spike emission, but the STBP propagates spike gradients across all neurons and time steps. Based on the recorded gradients $\partial \mathcal{L}/\partial S$, we have shown that our approaches successfully achieve sparse spike-driven propagation. Furthermore, we depict $\partial \mathcal{L}/\partial W$, that is the weight update $\Delta W$, in Fig.~\ref{fig:gradW}, where the distributions of weight update for the three algorithms are identical. Based on the recorded $\Delta W$, we have shown that our approaches can achieve the same learning effect as the STBP, even with sparse spike-driven propagation. Therefore, our approaches attain nearly identical learning effects to the well-established STBP but with lower computational costs.
\begin{table}[]
\centering
\caption{Training complexity of three algorithms}
\renewcommand{\arraystretch}{1.2}
\setlength{\tabcolsep}{6.5mm}
\begin{tabular}{ll}
\hline
Algorithms & \multicolumn{1}{c}{Complexity} \\
\hline
STBP    & $O({MNT}+{NT}+{NT})$ \\ 
MPD-ED  & $O({\zeta_{i}MNT}+{NT}+{\zeta_{o}NT})$ \\
STD-ED  & $O({\zeta_{i}MNT})$ \\ 

% STBP    & $O({\color[RGB]{255,0,255}MNT}+{\color[RGB]{0,176,80}NT}+{\color[RGB]{88,88,255}NT})$ \\ 
% MPD-ED  & $O({\color[RGB]{255,0,255}\zeta_{i}MNT}+{\color[RGB]{0,176,80}NT}+{\color[RGB]{88,88,255}\zeta_{o}NT})$ \\
% STD-ED  & $O({\color[RGB]{255,0,255}\zeta_{i}MNT})$ \\ 
\hline
\end{tabular}
\label{table:complexity}
% \vspace{-0.3cm}
\end{table}
% M-pre layer number N-post layer number

\begin{table}[]
\centering
\caption{Training complexity reduction of our methods in the second FC layer}
\renewcommand{\arraystretch}{1.4}
\setlength{\tabcolsep}{3.8pt}
\begin{tabular}{lllll}
\hline
\multirow{2}{*}{} 
&Training   & Mean activity   & Mean activity  & Complexity  \\ 
&algorithm  &  hidden layer & output layer & reduction \\ 
\hline
\multirow{2}{*}{N-MNIST} 
&STD-ED     &   0.0086      & 0.0604    & 99.14\%  \\
\cline{2-5}
&MPD-ED     &     0.0527    & 0.1287   &  94.22\% \\
% \multirow{2}{*}{} & Training & \multicolumn{1}{c}{Average} & \multicolumn{1}{c}{Average} & Complexity \\
%  & algorithm & \multicolumn{1}{c}{activity $\zeta_i$} & \multicolumn{1}{c}{activity $\zeta_o$} & reduction \\ \hline
% \multirow{2}{*}{N-MNIST} & STD-ED & 0.0086 & 0.0604 & 99.14\% \\ \cline{2-5} 
%  & MPD-ED & 0.0527 & 0.1287 & 94.22\% \\ 
\hline
\end{tabular}
\label{table:energy}
% \vspace{-0.3cm}
\end{table}
\subsection{Hardware implementation}

To demonstrate the energy efficiency and practicality of the proposed event-driven learning algorithm, we deploy the MPD-ED on a newly developed neuromorphic chip~\cite{zhang202322} to perform electromyography (EMG)--based hand gesture recognition.  The EMG-based gesture recognition plays an active role in various domains like human-machine interaction (HMI)~\cite{gordleeva2020real}, sign language interpretation~\cite{wu2016wearable}, healthcare~\cite{faust2018deep} and rehabilitation medicine~\cite{mcmanus2020analysis}. 
Due to their placement at the edge and reliance on body position for data collection, wearable EMG devices necessitate low power consumption and efficient on-chip learning capabilities. Since wearable EMG collection devices are placed at the edge and the collected signal varies across different body positions, computational models are required to have low power consumption and efficient on-chip learning abilities. In this experiment, we will demonstrate that the proposed event-driven learning approach for SNNs can efficiently fulfill these requirements.

\subsubsection{Dataset}We utilize the EMG hand gesture dataset provided by Ceolini et al.~\cite{ceolini2020hand}, which records the forearm muscle activity of participants through the EMG armband sensor Myo, as illustrated in Fig.~\ref{fig:setup}. The dataset is collected from 21 participants, with each conducting 3 trials. In each trial, participants perform five gestures: \textit{pinky}, \textit{elle}, \textit{yo}, \textit{index}, and \textit{thumb}, with each gesture repeated five times. The recording duration for each gesture is 2 seconds, interspersed with a 1-second relaxation interval between gestures. During the relaxation period, the muscle returns to the resting position, effectively eliminating any lingering activity from the preceding gesture. We split the collected data into training and testing sets in a 2:1 ratio. In addition, the neuromorphic chip processes spike events, which are generated based on differences in EMG signals between consecutive time steps. Each obtained spike event consists of 391 time steps, and we sample each spike event every 10 time steps. Consequently, each sample sent to the chip comprises 39 time steps.

\subsubsection{Hardware mapping} 
\begin{figure*}
\centering
\subfigure[]{\label{fig:setup}
\includegraphics[height=3.1cm]{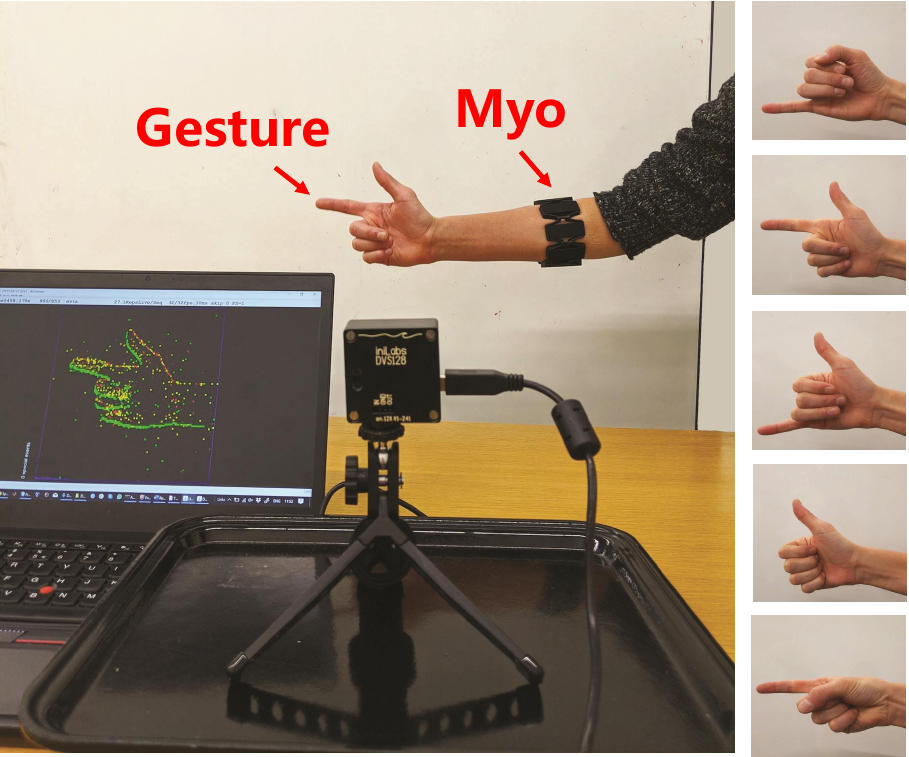}}
\hspace{-0.2cm}
\subfigure[]{\label{fig:diephoto}
\includegraphics[height=3.2cm]{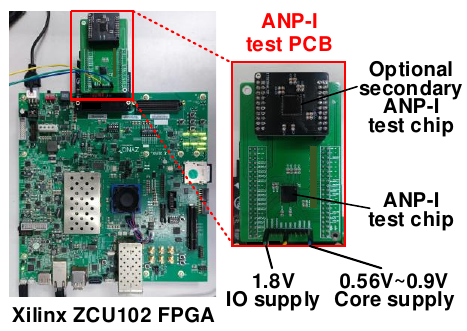}}
\hspace{-0.3cm}
\subfigure[]{\label{fig:ANP-I}
\includegraphics[height=3.3cm]{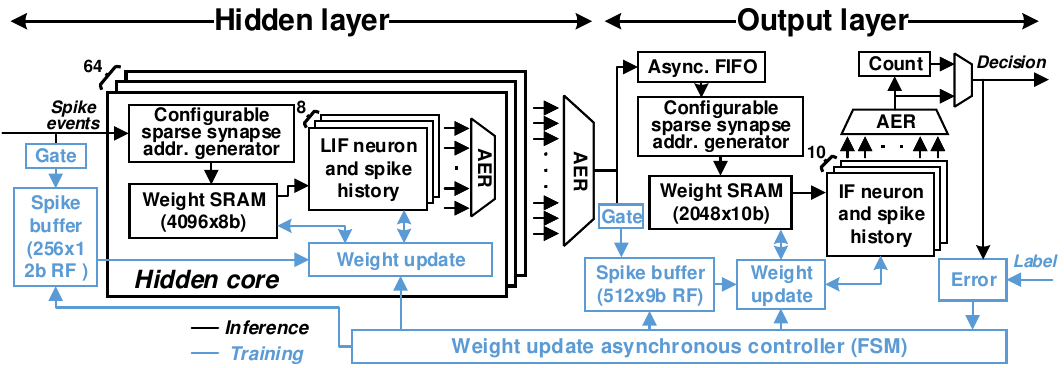}}
\caption{(a) Data collection setup and five gestures, which are \textit{pinky}, \textit{elle}, \textit{yo}, \textit{index}, and \textit{thumb}, respectively~\cite{ceolini2020hand}. (b) Measurement platform of the neuromorphic chip ANP-I~\cite{zhang202322}. (c) System diagram of the ANP-I~\cite{zhang202322}.}
\end{figure*}

We implement the MPD-ED on the ANP-I~\cite{zhang202322}, shown in Fig.~\ref{fig:diephoto}, a newly developed asynchronous neuromorphic chip with on-chip learning capability. The ANP-I implements the 1024-512-10 topology and performs neuronal dynamics in discrete time steps. As depicted in Fig.~\ref{fig:ANP-I}, it incorporates 522 neurons and 517K synapses on-chip, with a weight precision of 8/10 bits for each synapse. The threshold on ANP-I is fixed, so we cannot directly map the AFT-LIF neuron utilized in the MPD-ED onto it. To overcome this challenge, we emulate the adaptive firing threshold by increasing the membrane potential of non-firing neurons. Specifically, we introduce an auxiliary neuron in the presynaptic layer, and it is connected to all postsynaptic neurons with identical positive weights. After each time step simulation, the auxiliary neuron emits a spike, allowing us to add an identical value, i.e., the positive weight, to the membrane potential of non-firing neurons. This process occurs during both the training and inference stages. In the training process, to ensure that the weights of the auxiliary neuron remain unaltered, they are reset after each training iteration.

\subsubsection{Results}We feed spike events into the ANP-I chip for on-chip learning and assess the learning energy. The entire dataset undergoes training for 50 epochs, with all weights randomly initialized. We record the learning energy required for one sample during the training process, which is a pivotal metric in hardware assessments for gauging algorithmic energy efficiency. As the training progresses, the learning energy decreases due to the increasing sparsity of spike firing. Consequently, the average learning energy of the MPD-ED over the whole training phase is 312nJ/sample.

\begin{table}[]\small
\centering
\caption{The neuromorphic chip specifications and results}
\renewcommand{\arraystretch}{1.75}
\setlength{\tabcolsep}{8pt}
\begin{tabular}{|
>{\columncolor[HTML]{f1f1f1}}l |ll|}
\hline
Technology & \multicolumn{2}{l|}{28nm CMOS} \\ \hline
Core size & \multicolumn{2}{l|}{0.78$\times$1.63 mm$^2$} \\ \hline
On-chip memory & \multicolumn{2}{l|}{266.5kB SRAM} \\ \hline
Supply voltage & \multicolumn{2}{l|}{0.56V$\sim$0.9V} \\ \hline
Weight precision & \multicolumn{2}{l|}{8-bit (hidden), 10-bit (output)} \\ \hline
Power consumption & \multicolumn{2}{l|}{\begin{tabular}[c]{@{}l@{}}2.91mW@40MHz, 0.56V\\ 56.8mW@210MHz, 0.9V\end{tabular}} \\ \hline
Energy efficiency & \multicolumn{2}{l|}{\begin{tabular}[c]{@{}l@{}}1.49pJ/SOP@40MHz, 0.56V\\ 4.16pJ/SOP@210MHz, 0.9V\end{tabular}} \\ \hline
\cellcolor[HTML]{f1f1f1} & \multicolumn{1}{l|}{MPD-ED} & 312nJ \\ \cline{2-3} 
\multirow{-2}{*}{\cellcolor[HTML]{f1f1f1}\begin{tabular}[c]{@{}l@{}}On-chip learning\\ energy/sample\end{tabular}} & \multicolumn{1}{l|}{STBP} & 9320nJ \\ \hline
\end{tabular}
\label{tab:specifications}
\end{table}

In addition, we simulate the traditional surrogate gradient algorithm STBP on the ANP-I chip for energy comparison. The STBP conducts gradient backpropagation at every time step regardless of spike emission, so we activate each neuron to emulate the way STBP works. Employing identical input and hardware configurations, we record the learning energy required for one sample for the STBP throughout the training process. The average learning energy of the STBP over the whole training phase is 9320nJ/sample, nearly 30 times higher than that of MPD-ED. Finally, we summarize the neuromorphic chip specifications and results in Table ~\ref{tab:specifications}.

\section{Conclusion}
Due to the sparse event-driven nature, the advantages of SNNs in feedforward inference have been extensively investigated. However, how to effectively train
deep SNNs in an event-driven manner to reduce learning costs remains an open question. In this paper, we delve deeper into the sparse event-driven nature of SNNs in backward propagation, aiming to minimize the learning cost and maximize the energy-saving advantages of SNNs. Specifically, we propose two novel event-driven learning algorithms, namely STD-ED and MPD-ED, which leverage precise spike timing and membrane potential to perform event-driven backpropagation, respectively. The proposed STD-ED and MPD-ED methods achieve state-of-the-art accuracy performance, surpassing their counterparts by up to 2.51\% for STD-ED and 6.79\% for MPD-ED on the CIFAR-100 dataset. In addition, we theoretically demonstrate the significant energy efficiency of these proposed event-driven algorithms, where the STD-ED achieves a 99.14\% reduction in training complexity and MPD-ED achieves a 94.22\% reduction. More importantly, we successfully apply our work to a practical EMG-based hand gesture recognition task, strongly proving that our approach can meet the stringent requirements of edge devices for low power consumption and efficient on-chip learning capabilities. 

The event-driven learning algorithms proposed in this study, namely STD-ED and MPD-ED, are inspired by traditional backpropagation (BP) and backpropagation through time (BPTT) algorithms, which were originally designed for dense and analog ANNs.
However, the spikes in SNNs are sparse in both spatial and temporal domains, operating in an event-driven manner.
As a result, the backpropagation-based learning algorithms that are applicable in ANNs are not well-suited for sparse event-driven SNNs.
Furthermore, these backpropagation-based algorithms necessitate significant memory capacity and training resources for on-chip learning, which contradicts the limited resource characteristics of edge computing.
In our current and future work, we will focus on leveraging bio-plausible learning rules, such as Spike Timing Dependent Plasticity (STDP), to develop a local event-driven learning algorithm that can support the efficient implementation of low-power neuromorphic hardware for deeper SNNs.

\bibliographystyle{IEEEtran} 
\bibliography{ref}
\end{document}